\documentclass[journal]{IEEEtran}
\usepackage{cite,color,url}
\usepackage{amsmath,graphicx,amsfonts}
\ifCLASSINFOpdf
\else
\fi
\usepackage{amsmath,amssymb}
\usepackage{algorithm}
\usepackage{algorithmic}
\usepackage{stfloats}
\usepackage{multirow}
\usepackage{booktabs}
\usepackage{makecell}
\usepackage{threeparttable}

\usepackage{array}

\newcommand{\myfig}[1]{Fig.\;\ref{#1}}

\begin{document}

\title{Fast and Accurate FSA System Using ELBERT: \\An Efficient and Lightweight BERT}

\author{Siyuan~Lu,
        Chenchen~Zhou,
        Keli~Xie,
        Jun~Lin,~\IEEEmembership{Senior Member,~IEEE},
        and~Zhongfeng~Wang,~\IEEEmembership{Fellow,~IEEE}\vspace{-1em}
\thanks{This paper is an extended version of our conference paper\cite{xie2021elbert} which has been presented at the IEEE International Conference on Acoustics, Speech and Signal Processing (ICASSP). This work was supported in part by the National Natural Science Foundation of China under Grant 62174084, 62104097, and in part by the High-Level Personnel Project of Jiangsu Province under Grant JSSCBS20210034. (Corresponding authors: Jun Lin; Zhongfeng Wang.) \IEEEcompsocthanksitem The authors are with the School of Electronic Science and Engineering, Nanjing University, Nanjing 210023, China (e-mail: \{sylu,cczhou,klxie\}@smail.nju.edu.cn, \{jlin, zfwang\}@nju.edu.cn)}
}

\markboth{Journal of \LaTeX\ Class Files}
{Shell \MakeLowercase{\textit{et al.}}: Bare Demo of IEEEtran.cls for IEEE Journals}

\maketitle

\begin{abstract}
With the development of deep learning and Transformer-based pre-trained models like BERT, the accuracy of many NLP tasks has been dramatically improved. However, the large number of parameters and computations also pose challenges for their deployment.
For instance, using BERT can improve the predictions in the financial sentiment analysis (FSA) task but slow it down, where speed and accuracy are equally important in terms of profits.
To address these issues, we first propose an efficient and lightweight BERT (ELBERT) along with a novel confidence-window-based (CWB) early exit mechanism. Based on ELBERT, an innovative method to accelerate text processing on the GPU platform is developed, solving the difficult problem of making the early exit mechanism work more effectively with a large input batch size.
Afterward, a fast and high-accuracy FSA system is built.
Experimental results show that the proposed CWB early exit mechanism achieves significantly higher accuracy than existing early exit methods on BERT under the same computation cost.
By using this acceleration method, our FSA system can boost the processing speed by nearly 40 times to over 1000 texts per second with sufficient accuracy, which is nearly twice as fast as FastBERT, thus providing a more powerful text processing capability for modern trading systems.
%








\end{abstract}


\IEEEpeerreviewmaketitle

\section{Introduction}
\IEEEPARstart{S}{entiment} analysis is a branch of natural language processing (NLP) and attracting more and more attention from the financial area.
Aiming to identify the sentiment from text such as news, analyst commentary, and social media, financial sentiment analysis (FSA) has been widely used in company assessments, report generation, and trading strategies.
In this article, we are mainly concerned with the FSA system used for the quantitative or event-driven investment field, where the accuracy and the processing speed are both important.

Nowadays, more and more quantitative investors are pouring into the market, and quantitative funds are also managing more and more money. Along with this trend, the profit margins of many conventional factors are getting smaller and smaller, and investors have to turn their attention to higher-dimensional data to obtain higher excess returns.
Fig. 1 describes such a modern quantitative investment system where the so-called alternative data are used. 
Among these alternative data, the most useful are still various text data, which are much easier to obtain from the Internet and have also been long relied on by investors to make buying and selling decisions.
Therefore, with the ability to process and mine large amounts of financial text data, FSA becomes an essential module in this system.

Although many research FSA works have been done, there are still problems if we want to apply them to the industry. Simple models such as dictionaries \cite{davis2015effect,zhang2010trading} and SVMs \cite{smailovic2014stream} often have insufficient accuracy, while large deep neural networks can give more precise results. However, the computation of these networks, especially the NLP Transformers \cite{mishev2020evaluation} is more time-consuming and will significantly increase the delay of the whole investment system.

Starting from the actual demand of FSA and exploring model compression for BERT (the most commonly used NLP Transformer for text classification), we introduce a fast and lightweight BERT named ELBERT and develop a high-throughput and high-accuracy FSA system.
The ELBERT uses the parameter sharing strategy from ALBERT\cite{lan2019albert} and is further improved by our proposed confidence-window-based (CWB) early exit mechanism. The early exit is a very practical acceleration method and suitable for Transformer neural networks \cite{liu2020fastbert}.
However, the efficiency of existing early exit methods will be influenced if the network has a large batch size and is deployed on widely-used GPU platforms (shown in Fig. \ref{fig:ELBERTFAST}). Usually, early exit works more effectively only when the inference batch size equals 1. 
To resolve this issue, we also design an efficient acceleration method based on ELBERT, which combines the early exit mechanism artfully with batch-level parallel computing.
Experimental results demonstrate that during inference, ELBERT is able to achieve an adaptive speedup varying from 2$\times$ to 10$\times$ with negligible accuracy loss in many NLP tasks. Furthermore, by using our acceleration method, the FSA system acquires a speedup of nearly 40$\times$ without operator or CUDA level optimizations.

Our proposed datasets and the code of ELBERT (using pytorch) are publicly released on Github, and main contributions include the following:


\begin{itemize}
\item We introduce a fast and light ELBERT\footnote{\footnotesize{\url{https://github.com/shakeley/ELBERT}}} model, which has fewer parameters (18M) and uses our proposed CWB early exit mechanism to accelerate the inference. On many NLP tasks (not only FSA), the CWB mechanism outperforms existing early exit methods to accelerate BERT under the same computation cost. 
\item Based on ELBERT, an efficient text processing acceleration method is developed to solve the contradiction between large input batch size and the early exit mechanism.
With this method, the model can avoid the problem of computational efficiency degradation caused by samples of different difficulty exiting at different layers.
As far as we know, this is the first method among open literature that can do this.

\item A high-speed and high-accuracy FSA system (implemented with an Nvidia RTX 3090 GPU) is built based on the ELBERT and the acceleration method. With sufficient classification accuracy, we achieved throughputs of 1503 text/s and 1107 text/s on two FSA datasets, respectively.
\item We propose a new FSA dataset labeled by our experts, which is also available in another Github project\footnote{\footnotesize{https://github.com/NLP-Applications/Financial-sentiment-analysis-NLPTransformers.git}} and evaluated in our experiments.
\end{itemize}

  \begin{figure*}[t]
    \centering
    \includegraphics[width=17cm]{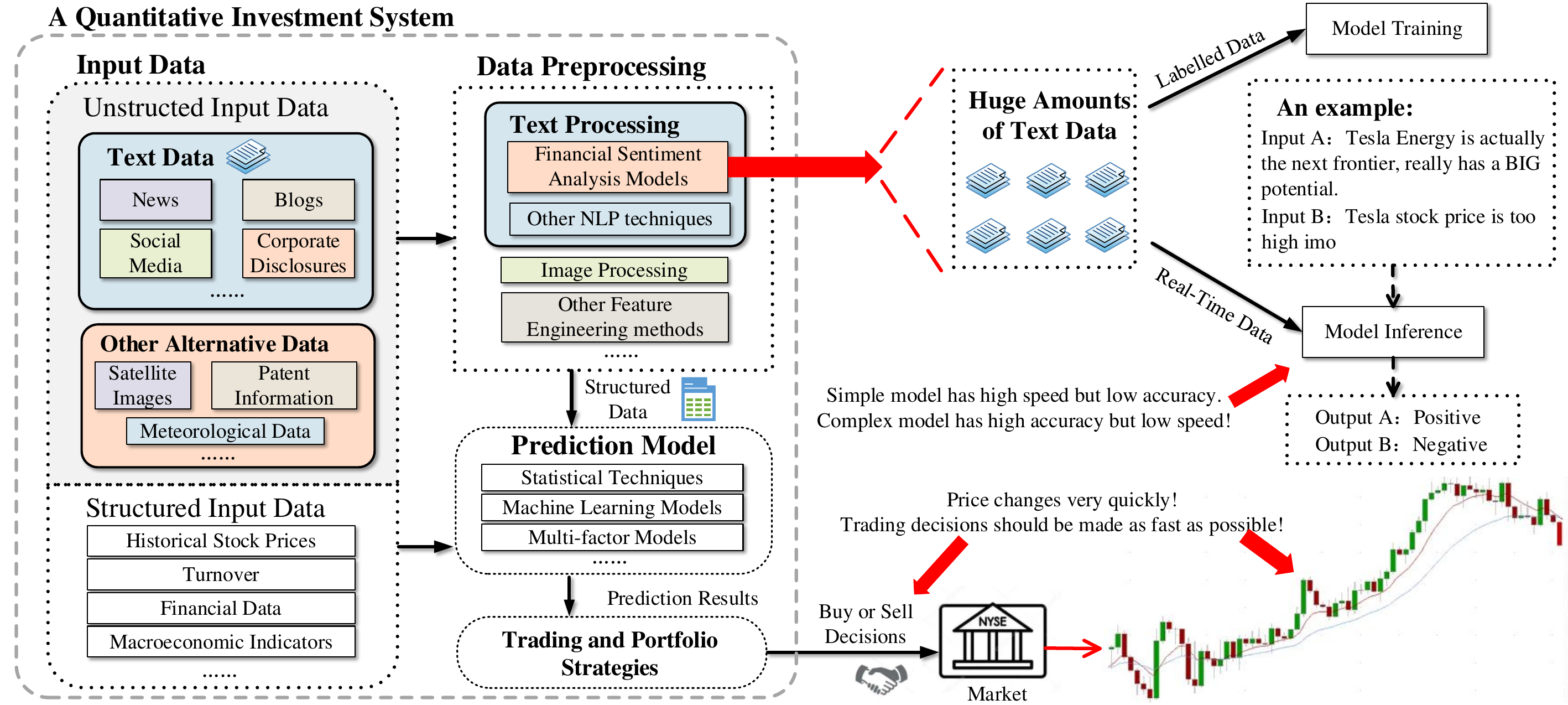}
    \caption{The structure of a modern quantitative investment system where FSA models are used. This figure also explains why we need a fast and high-accuracy FSA model. The example in it is from \cite{sawhney2021fast}.
    }
    \label{fig:BERT2ELBERT}
  \end{figure*}

The rest of this paper is organized as follows. Section \uppercase\expandafter{\romannumeral2} gives a review of the background and related works.
The ELBERT model is presented in Section \uppercase\expandafter{\romannumeral3}.
In Section \uppercase\expandafter{\romannumeral4}, we introduce the FSA system along with the ELBERT-based acceleration method on GPU and the proposed datasets.
Experimental results are given in Section \ref{sec:experiments1} and Section \ref{sec:experiments2}.
Section \ref{conclusion} concludes this paper.

\section{Background and Related Works}

In this section, we will first introduce the Transformer model and Transformer-based pre-trained language models, which are the backbone of our FSA system, and then give an overview of related works, which are mainly distributed within two domains, including BERT compression and FSA. 
However, our work is far beyond simply applying existing model compression methods to the FSA task. 
Compared to existing works on BERT compression, a major advantage of our work is that it solves the problem of computational efficiency degradation of the early exit method when the batch size increases. 
Another advantage over existing works on FSA is that we give more consideration to its speed, which has been ignored by most related works. Although Mishev \textit{et al.} \cite{mishev2020evaluation} also compared the performance of different compressed BERT models on FSA datasets, it also does not discuss the speed or latency of these models. The following are the details.

\subsection{The Transformer and Transformer-Based Pre-Trained Models}
\begin{figure}[]
    \centering
    \includegraphics[scale=0.35]{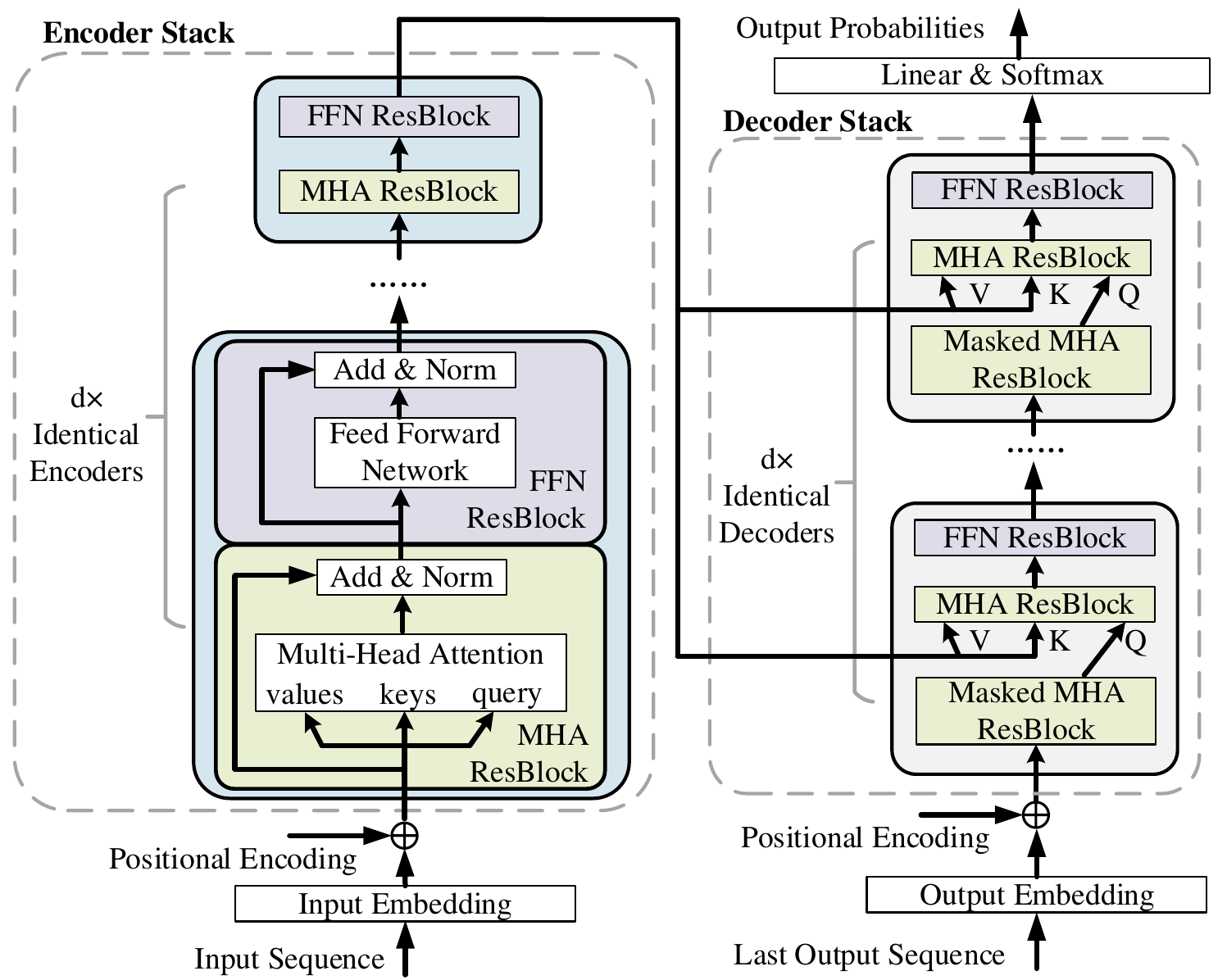}
    \caption{The model architecture of the Transformer. The depth of the network is denoted as d. The multi-head attention has three input tensors, including query (Q), keys (K), and values (V). Notice that in the encoder, Q, K, and V are equal, so the input tensor of encoder layer can be seen as one single tensor, the shape of which is (batch\_size, sequence\_length, hidden\_state).}
    \label{fig:Transmodel}
\end{figure}

Since the Transformer model (Fig.~\ref{fig:Transmodel}) \cite{vaswani2017attention} was invented in 2017, it has achieved great success and been widely used in the NLP area. Compared with the recurrent layer in recurrent neural networks (RNNs), the multi-head attention layer can not only learn long-range dependencies between words in the sequence, but also be trained in parallel. Recently, researchers find the Transformer model even outperforms the convolutional neural network (CNN) in multiple computer vision (CV) tasks \cite{liu2021swin}, which also proves the advancement of the Transformer model and the attention mechanism.

Based on the Transformer neural network architecture, many NLP pre-trained models (PTMs) have been proposed, making NLP techniques enter a new area. These NLP Transformers have been pre-trained on large corpora and can be finetuned for downstream tasks with little labeled data. After finetuning, they can achieve better performances than previous methods. A representative model among these PTMs is BERT (Bidirectional Encoder Representations from Transformers), which achieved state-of-the-art performance on eleven NLP tasks \cite{devlin2019bert}.
Since BERT was released in 2018, many BERT-based or Transformer-based PTMs have also been proposed in recent years, such as RoBERTa \cite{liu2019roberta}, ERNIE \cite{zhang2019ernie}, and XLNet \cite{yang2019xlnet}. These models further improved the accuracy of many NLP tasks. Meanwhile, PTMs are also getting bigger and bigger.
In the past two years, large models with trillions of parameters are popping up around the world, including GPT-3 \cite{floridi2020gpt}, Switch Transformer \cite{fedus2021switch}, and M6 \cite{lin2021m6}.

\subsection{NLP Transformers Compression}
\label{BackgroundCompress}
Owing to massive data used for training and extremely large numbers of trainable parameters, these big models can do many tasks like reading, writing, designing, and Q\&A (question and answer). However, applying these models in real applications faces big challenges caused by huge memory consumption and computational delay.
Apart from the accuracy or support for multitasking, it is also imperative to consider the cost of computing devices in the industry.
Therefore, the model compression of NLP Transformers has also become an important research field.
We divide prior works in compressing BERT into two categories:
\subsubsection{Structure-wise} Structure-wise approaches try to withdraw the trivial elements in networks. There are three common structure-wise compressing methods for BERT: weight pruning, quantization, and knowledge distilling. For weight pruning, Gordon \emph{et al.}\cite{gordon2020compressing} involved the magnitude-based pruning method to BERT, while Michel \emph{et al.}\cite{michel2019sixteen} pruned BERT based on gradients of weights. For quantization, Q-BERT\cite{shen2020q} employed a Hessian-based mix-precision approach to compress BERT, and Q8BERT \cite{zafrir2019q8bert} quantized BERT using symmetric linear quantization. In addition, knowledge distilling is applied by Tang \emph{et al.} \cite{tang2019distilling}, Sun \emph{et al.}\cite{sun2019patient}, DistillBERT\cite{sanh2019distilbert} and TinyBERT\cite{jiao2019tinybert} to build lightweight BERTs. Compared with the compressed models based on these techniques, ALBERT \cite{lan2019albert} drastically decreases the number of parameters and storage consumption by parameter-sharing strategy and even outperforms BERT.
\subsubsection{Input-wise} Input-wise methods avoid superfluous calculations depending on the different complexness of inputs. For example, BranchyNet\cite{teerapittayanon2016branchynet} proposed the entropy-based confidence measurement, and Shallow-Deep Nets\cite{kaya2019shallow} solved the overthinking issue with an early exit mechanism. DeeBERT\cite{xin2020deebert} and TheRT\cite{Schwartz2020TheRT} applied the basic early exit method to BERT. FastBERT\cite{liu2020fastbert} proposed a self-distilling method in fine-tuning. Unlike these works that only focus on the confidence of the classifier, our CWB mechanism also incorporates the historical trend of the classifier prediction.

It is worth mentioning that this classification method for these prior works is also beneficial for us to get the idea of ELBERT, which will be further discussed in Section \ref{ELBERT}. From another point, due to the requirements of speed and accuracy, the FSA is also an excellent area to study model compression of BERT, and the details are provided as follows.

\subsection{FSA}

Different from the general sentiment analysis, which may be ineffective when applied to particular domains like finance, FSA aims at guessing the reaction of financial markets to the information extracted from news, corporate announcements, or social media.
With the results of FSA, which are usually positive or negative, investors and traders can adjust their decisions in advance \cite{mishev2020evaluation}.
Previous works have shown the effectiveness of applying machine learning techniques in FSA tasks. \cite{smailovic2014stream} employed the Support Vector Machine (SVM) classifier to sentiment analysis of financial tweet streams and showed the relationship between the positive sentiment and the changes in stock closing prices. \cite{kraus2017decision} predicted stock-market movements by applying Long short-term memory (LSTM) networks to company announcements, and showed the deep learning method is preciser than conventional approaches. However, compared with NLP Transformers, these methods are not accurate enough \cite{mishev2020evaluation}.

As a commonly used text feature extractor in various AI applications, the BERT or Transformer encoder layer is also applied in the FSA task and show promising results. The FinBERT\cite{araci2019finbert} is a BERT-based model designed to tackle NLP tasks in financial domain, and achieved the state-of-the-art results on two FSA datasets. \cite{mishev2020evaluation} studied a lot of NLP-based methods for FSA, and drew a conclusion that NLP Transformers show superior performances compared to various sentiment analysis approaches. To mine the key information from online financial text and public opinion analysis in social media, \cite{zhao2021bert} proposed a sentiment analysis and key entity detection approach based on BERT.
Additionally, as an important application of FSA, the event-driven trading strategy uses the results of classifying financial texts and make buying or selling decisions \cite{fataliyev2021stock}. Recently, many latest event-driven trading studies \cite{sawhney2021quantitative,zhou2021trade,wu2020event} also utilize BERT or Transformer encoder layers to help increase the accuracy of text classification and the profitability.
These works prove that BERT or Transformers can significantly improve the accuracy of FSA. However, the problem of time consumption of these novel networks has not received as much attention.

As a text-processing part of a quantitative investment system or event-driven trading system, which requires extremely short processing latency (millisecond or even microsecond level) to reduce slippage, the speed of a FSA system is a really important performance indicator. On one hand, when an event which has a significant impact on the market occurs, the time left for us to invest is limited. Financial experts agree that the time for the market to absorb news impact is indeterminate, and usually ranges from minutes to hours \cite{li2021modeling}.
On the other hand, to increase the profitability as much profit as possible, the FSA system should finish processing every single input sentence as quickly as possible.
In the modern times of information explosion, news and opinions from Internet are being produced more and more frequently with the development of social media. This situation can provide really good investment opportunities, only if we capture the granularity of online text posting times and make better and faster responses than other traders to market changes \cite{sawhney2021fast}.
Hence, a high-speed and high-accuracy FSA system is desired so that we can finish processing as much text data in the same amount of time as possible, or a fixed number of data in less time.

Several FSA and event-driven trading works mentioned the importance of speed. However, these works all failed to discuss how to apply model compression or acceleration techniques to improve text processing speed.
For example, authors of \cite{zhou2021trade}
pointed out that the profitability of an event-driven model highly relies on how ``quick'' it can perform tradings after a piece of news is posted. However, the speed of their Transformer-based detecting model was not discussed. In \cite{mishev2020evaluation}, some lightweight BERT models were fine-tuned for FSA to compare their accuracy, but each model's processing speed was not measured either.

Although the NLP Transformers have been applied in FSA and achieved good results in accuracy, using more extensive networks may bring higher latency and bigger slippage, which is ignored by existing works. Can we hold the high accuracy and boost the speed of FSA at the same time? Using some very ingenious model compression techniques, we will try to answer this question in this paper.

\section{ELBERT}
\label{ELBERT}
This section describes the proposed CWB early exit mechanism and ELBERT to show our contribution to model compression of BERT. Here we focus on the network architecture and the computation flow of ELBERT. To understand the specific operations for utilizing the ELBERT model to accelerate the text processing and the advantages significantly, please refer to Section \ref{ssec:accELBERT}.

  \begin{figure}[t]
    \centering
    \includegraphics[width=8.5cm]{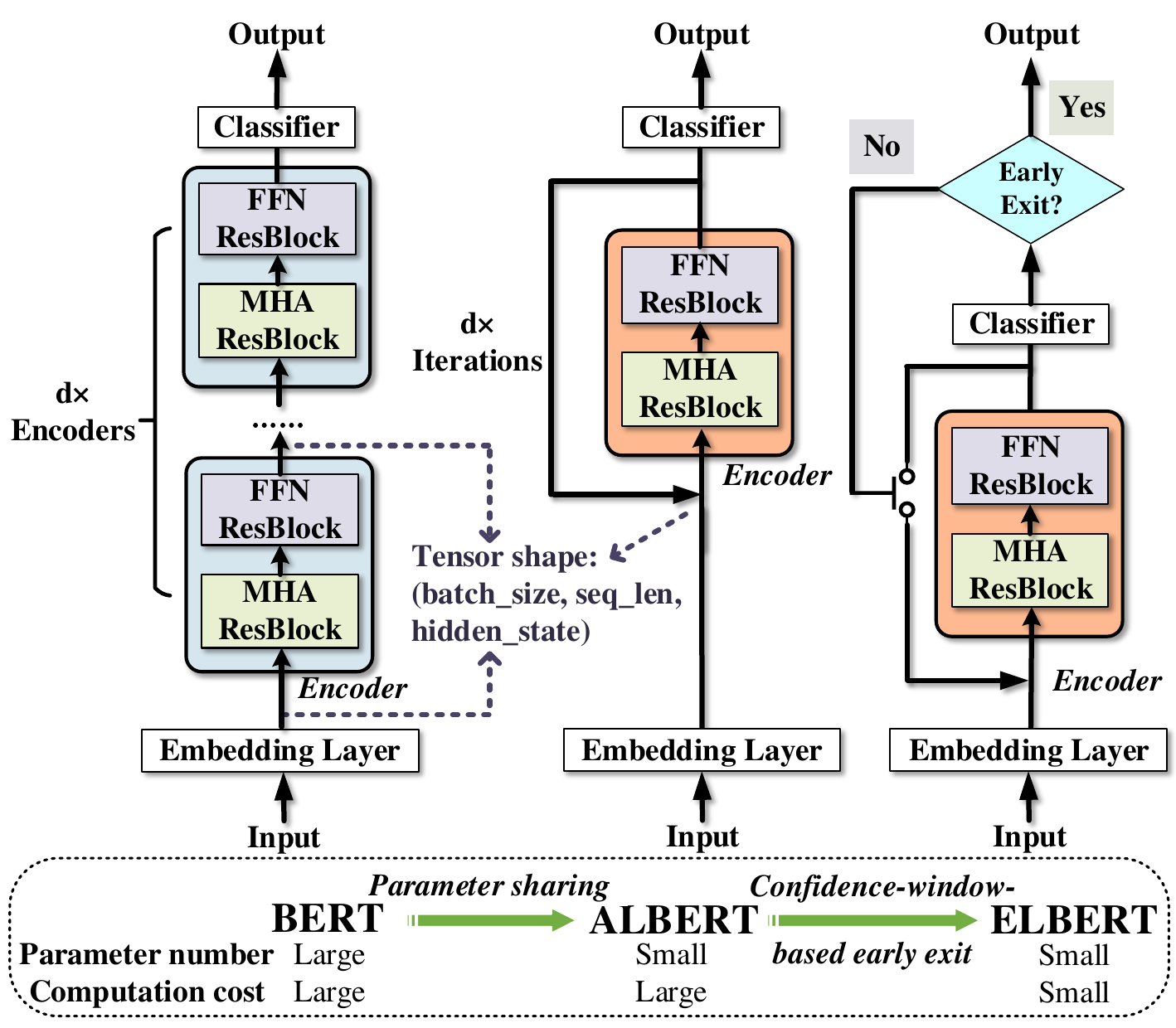}
    \caption{ALBERT adjusts from BERT, and ELBERT changes from ALBERT. Compared with ALBERT, ELBERT has no more extra parameters, and the calculation amount of early exit mechanism is also ignorable\cite{liu2020fastbert} (less than 2\% of one encoder).
    }
    \label{fig:BERT2ELBERT}
  \end{figure}
\subsection{Overview}

By making all the encoder layers use the same parameters, the ALBERT dramatically reduces the number of parameters, and it can still achieve competitive accuracy results.
However, this model can not reduce any computational delay.
Using ALBERT as the backbone model, we propose the ELBERT, the architecture of which is described in \myfig{fig:BERT2ELBERT}.
Owing to the CWB early exit mechanism, the average inference speed of ELBERT is improved significantly compared with ALBERT, and no additional parameters or training overhead is introduced.
By this means, we combine the structure-wise and input-wise compression methods to sufficiently reduce redundant computations so that a lightweight model can be obtained, which has fewer parameters and adjustable computing time.
Training and inference of ELBERT are introduced next.

\subsection{Training}
To accommodate the early exit mechanism in inference, ELBERT calculates the losses of inputs exiting at different depths during training.
Note that the loss of prediction using each layer's hidden state $\mathbf{h_i}$.
For classification, the $i$-th layer $\mathcal{L}_i$ calculate the early exit loss with \emph{Cross-Entropy}:

\begin{equation}
  \mathcal{L}_{i}=-\sum_{c \in C}\left[ \mathbb{I} \left[\mathbf{\hat y}_{i}=c\right] \cdot \log P\left(\mathbf{\hat y}_{i}=c \mid \mathbf{h}_{i}\right)\right].
  \end{equation}

Here $c$ denotes one class label, and $C$ represents the set of class labels.
The conventional method is simply adding up $\mathcal{L}_i$ as the total loss $\mathcal{L}$ \cite{xin2020deebert,liu2020fastbert}.
To get better training results under various combinations of losses, we assign a trainable variable $t_i$ with an initial value of 4 to each layer, inspired by \cite{wang2019dynexit} and \cite{wang2021rethinking}.
Thus the weights of each layer's loss are dynamic, which is helpful to cover different cases.
The weight of the $i$-th layer $w_i$ is calculated by

\begin{equation} \label{weight}
  w_i=\left\{\begin{array}{ll}\sigma( t_{i}) & 0<i\le d-1 \\
  d-\sum_{i=1}^{d-1} \sigma( t_{i}) & i=d\end{array}\right. .
  \end{equation}

Note that we use $\sigma(\cdot)$ refers to sigmoid function $\sigma\left(t_{i}\right)=1 /\left(1+\exp({-t_{i})}\right)$, and the depth of ELBERT is defined as $d$.
Afterwards, we can calculate the total loss $\mathcal{L}$ with a weighted sum

\begin{equation}
    \mathcal{L}=\sum_{i=1}^{d}  w_{i}\cdot \mathcal{L}_{i}.
  \end{equation}

By this means, the cases where the input may drop out at different encoder layers can be well regarded, which is very helpful in bridging the gap between training and inference for ELBERT.

\subsection{Inference}
  \label{ssec:inference}

The CWB mechanism in the ELBERT has two stages.
This mechanism pays attention to the classifier output's intermediate state and historical trend to decide whether the calculation should be stopped early.
To help understand this, imagine taking an exam. When the answer is convinced to be correct, you can skip further checking to save time. Another situation is that when several checking results make you believe more and more in a certain option, there is also no need to continue calculating.

In short, in the first stage the model will decide whether to exit earlier based on whether the exit threshold is reached. 
If the threshold is reached, the computation of the input sample will stop early, otherwise, we enter the second stage.
In the second stage, the historical trend of the classifier will be further considered, and when this trend stabilizes, the exit condition is also satisfied to save calculation time. The following is a detailed description: 

During inference, firstly the input sentence $\mathbf{x}$ is given to the embedding layer, and the output of this layer is $\mathbf{h}_0$. Afterward, the hidden state tensor $\mathbf{h}_i$ will go through the encoder iteratively:

  \begin{equation}
    \mathbf{h}_i=\left\{\begin{array}{ll}
        Encoder(\mathbf{h}_{i-1}) & 0<i\le d \\
        Embedding(\mathbf{x}) & i=0\end{array}\right..
  \end{equation}

After each iteration, $\mathbf{h}_i$ will be utilized as the input of the classifier, and the output is a prediction probability distribution $p_i=Classifier(\mathbf{h}_i)$ computed by a fully-connected (FC) layer and softmax function for classification. Subsequently, the predicted label can be calculated as $\mathbf{\hat y}_i = argmax(p_i)$.

The first stage of the CWB mechanism focuses on the classifier's confidence or intermediate state. Given a probability distribution $p_i$, the normalized entropy is taken as the $Puzzlement$ of the current classifier:

  \begin{equation}
  Puzzlement(i) = \frac{\sum_{j=1}^{C} p_i(j) \log p_i(j)}{\log (1/C)}.
  \end{equation}

$C$ expresses the number of labeled classes.
When $Puzzlement(i) < {\delta}$, where $\delta$ is a user-defined threshold, the model will stop the inference in advance and take $\mathbf{\hat y}_i$ as the final prediction to skip further computations.
Therefore, if a faster model is needed and some accuracy degradation is tolerable, we can set a higher $\delta$, and vice versa.

The second stage tracks the historical trend of the classifier output over a time window, the size $W$ of which is defined according to user needs.
Please note that an appropriate size needs to be chosen for $W$. A too small $W$ may make this trend not stable enough, and a too large $W$ will prevent the model from achieving significant acceleration.
To seize the best moment to exit, we propose three criteria for triggering the second stage early exit \emph{in a time window}:
\begin{itemize}
 \item{The prediction probability distribution $p_i$ is more and more biased towards a certain class in $W$ times;}
  \item{The variation range of $p_i$ is less than a set value in $W$ times;}
  \item{The predicted label $y_i$ stays the same in $W$ times.}
  \end{itemize}

Experimental results show that the first criterion outperforms others. In subsequent experiments, we will use the first criterion for the second stage by default. 
We choose the best value of $W$, which is 8 and has the best performance on the validation set. Of course, we also recommend readers to take the setting of $W=8$ when using ELBERT, or try a little fine tuning on this number.

Usually, the moment when we get enough confidence is preferred. Only when the first stage condition is not satisfied will we consider the second stage early exit. By this means, the possibility of early exit is well covered, and the overthinking problem is mitigated, reducing the model's attention to some misleading information.
Note that to achieve different accuracy-speed tradeoffs in inference, we use a same trained model, merely adjusting one parameter without retraining.

\section{ELBERT-Based FSA System}
\label{section4}

\begin{figure}[t]
    \centering
    \includegraphics[width=8.5cm]{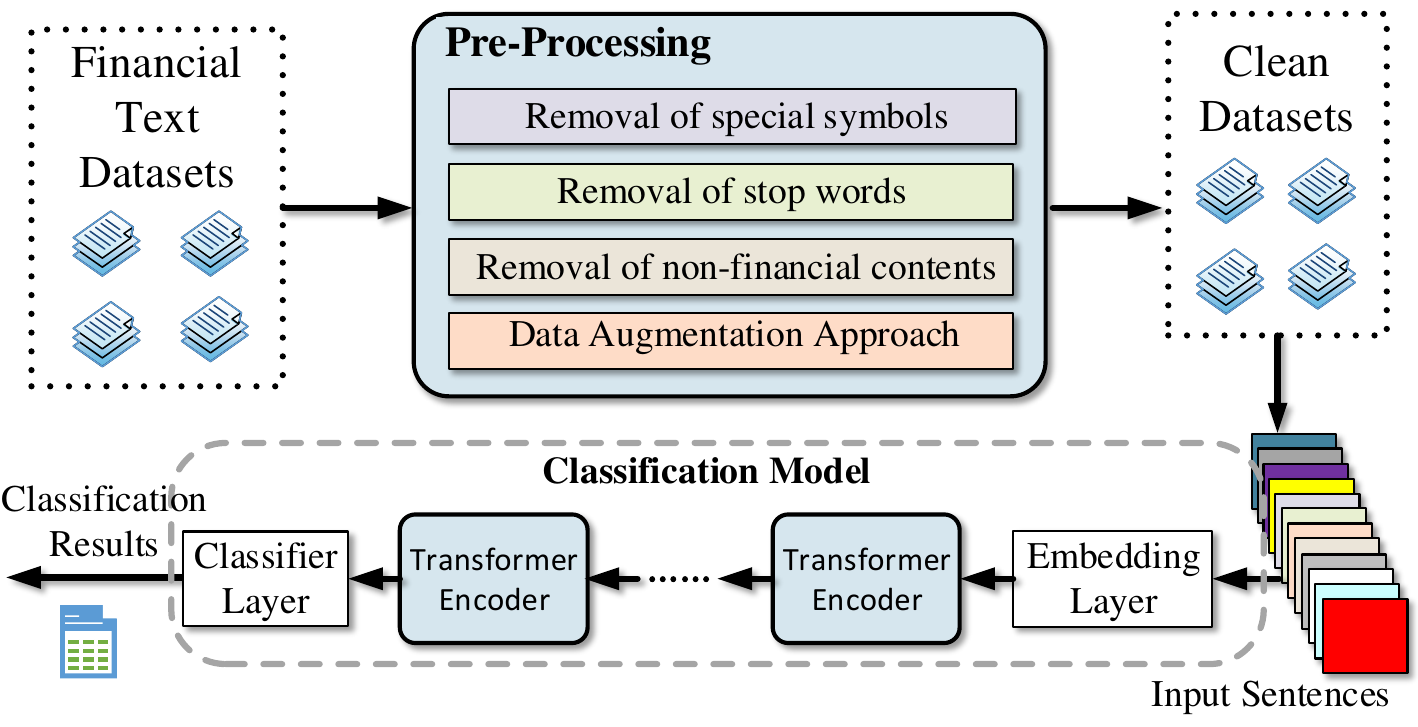}
    \caption{The top-level architecture of our FSA system.}
    \label{FSAoverview}
  \end{figure}

\begin{figure*}[t]
    \centering
    \includegraphics[width=17cm]{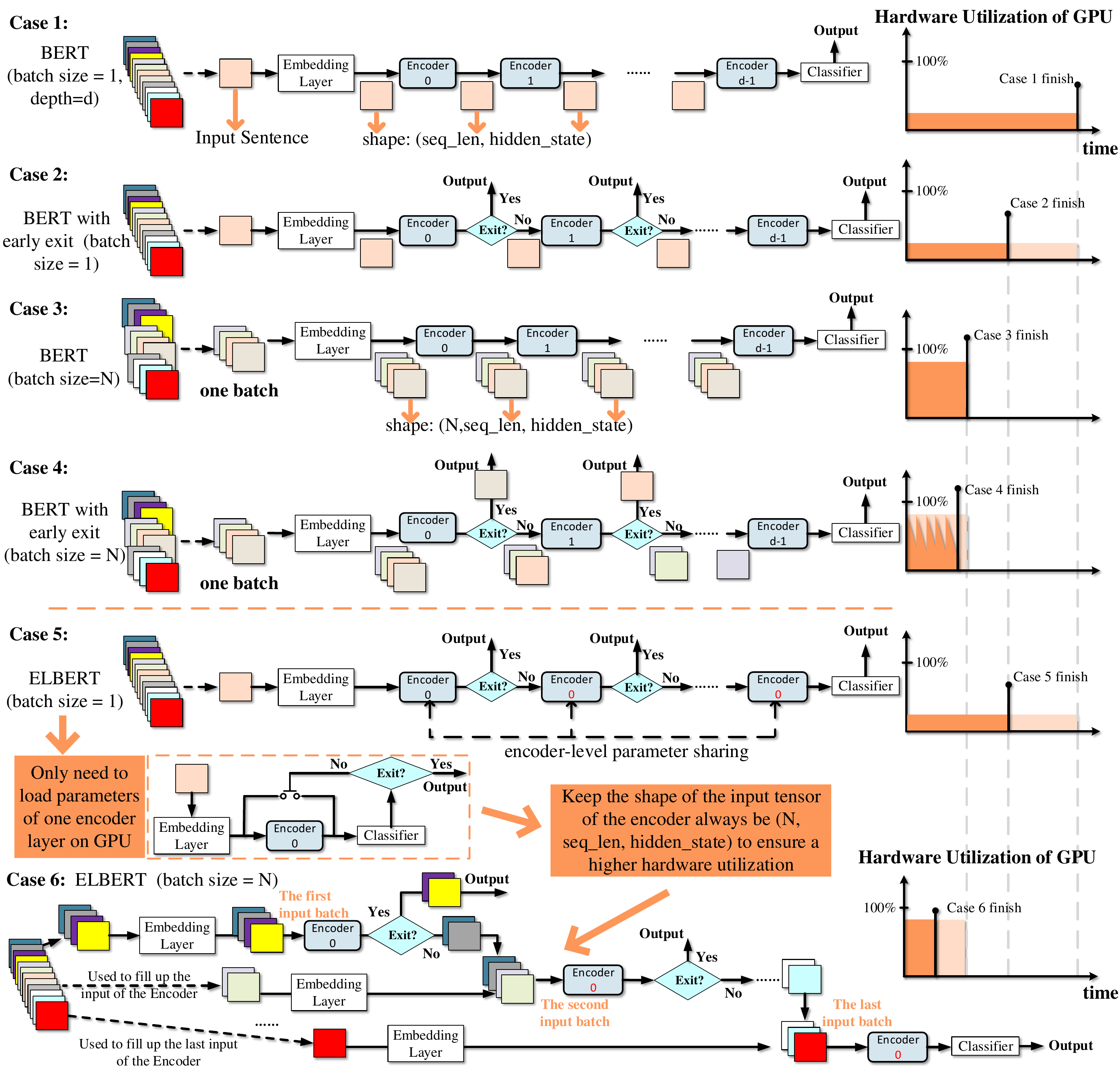}
    \caption{Several cases to explain the advantages and the basic idea of the proposed acceleration mechanism based on ELBERT.}
    \label{fig:ELBERTFAST}
  \end{figure*}

\subsection{Overview}

Fig. \ref{FSAoverview} briefly introduces the FSA system we used in this work. Before the classification model, texts need to be preprocessed to remove useless information. We take a part of the data pre-processing process in the previous paper\cite{usmani2021news}, which is a common method, and make some adjustments to suit our needs. Our pre-processing process includes: 1) Removal of special symbols: There are various symbols in the texts such as full stop, quotation marks, and ellipsis. These symbols do not contain sentiment. 2) Removal of stop words: Stop words add no meanings to a sentence. Conjunctions and determiners are parts of stop words. 3) Removal of non-financial contents: Some texts contain contents that are unrelated to finance. We also remove news sources and web links that have little relationship with financial emotions. 4) Data augmentation approach: The back-translation \cite{li2022data} is utilized in our system as a data augmentation approach.

Afterward, the NLP Transformers, which have been pre-trained on large corpora and will be fine-tuned on FSA datasets during the training stage, are chosen as the classification model.
Notice that the Transformer networks account for the vast majority of the system delay during the inference phase.
In particular, if the ELBERT is used as the classification model here, we can use a clever method to improve the inference speed, which will be discussed in Section \ref{ssec:accELBERT} in greater detail.

\subsection{Inference Acceleration of ELBERT on GPU}
\label{ssec:accELBERT}

Servers with GPU have become very general computing devices, widely used in artificial intelligence and cloud computing, and have been equipped by many quantitative investment institutions.
Here we give an innovative and inspiring method to accelerate BERT on GPU based on the proposed ELBERT. In this method, the early exit mechanism is artfully combined with batch-level parallel computing, which would be a quite difficult problem before we invented the ELBERT.





In Fig.~\ref{fig:ELBERTFAST}, we give several cases of accelerating BERT on GPU to explain the advantages and the basic idea of the proposed acceleration mechanism. The relationship between the GPU hardware utilization and the computational delay of each case is also roughly described on the right side of the graph. The following is a detailed introduction to the description of this figure:

\subsubsection{Case 1} First of all, Case 1 is the simplest one, without using any acceleration techniques and having a batch size equal to 1. The input sentence will first pass through the embedding layer. The output tensor of the embedding layer has a shape like (seq\_len, hidden\_state), where ``seq\_len'' denotes the maximum value of the sequence length of all the input sentences, and ``hidden\_state'' means the dimension of the model (which is the $d_{model}$ in \cite{vaswani2017attention}). Notice that the input tensor and the output tensor of a Transformer encoder have the same shape, so the inputs and outputs of all the encoders in Case 1 will be shaped as (seq\_len, hidden\_state).
\subsubsection{Case 2} In Case 2, once the exit condition is reached, the calculation in an intermediate layer of BERT will be terminated in advance, thereby avoiding redundant operations and achieving the purpose of acceleration. Compared with Case 1, the speedup of Case 2 depends on how many layers the network skips. However, this strategy will not improve the hardware utilization of GPU, which is the most commonly used hardware coprocessor for deep learning.
\subsubsection{Case 3} Case 3 introduces another frequently used method to accelerate neural network inference. Actually, when the GPU is used to accelerate the inference (this phenomenon also occurs in the training phase) of a neural network, it is found that if the batch size is set to 1, the utilization of the computation or memory resources of the GPU tends to be very low. Therefore, a straightforward method would be increasing the batch size to improve the parallelism of computation so that the overall latency will be reduced. If we enlarge the batch size under a specific limit, the throughput of the inference system will be enlarged by a nearly equal proportion. 
Of course, this approach does not reduce the computational load of the network itself as early exit does.
\subsubsection{Case 4} Compared with model compression techniques such as quantization and pruning, which often need dedicated hardware to release their abilities to accelerate the network, the early exit does not depend on specific computing devices. However, it is not well compatible with increasing the batch size, which is illustrated in Case 4. Different samples in one batch will exit at different network layers, making the latency of the entire network be decided by the slowest sample of every single batch, which could be described as a ``Wooden barrel effect'' (if there is a short slab, the barrel will leak). Therefore, only if the batch size is one the early exit method can obtain a better speed-up.
\subsubsection{Case 5 and Case 6} As a matter of fact, the proposed ELBERT model is an excellent solution to the problem of the incompatibility between early exit and increasing batch size. Due to the encoder-level parameter sharing mechanism, the GPU will only need to load one encoder layer, and the computation between the embedding layer and the classifier will be like a repetitive iteration process. For example, although input sentence A and input sentence B are not sent into the network simultaneously (maybe A is going to start its fifth iteration, but B has just passed through the embedding layer), they can still be computed in a single batch. By adjusting the input sequence, we can fill up every input tensor of the encoder during most time of the whole computation (except only a few iterations near the end) to make its shape always be like (N, seq\_len, hidden\_state), so that the high utilization of GPU will be ensured and the text processing speed can be further improved.


The detailed computational flow of the acceleration method is given in Algorithm 1.
$\hat{S}$ represents the input tensor of the encoder, and its shape is (N, seq\_len, hidden\_state). $\hat{F}_i$ is used to mark whether the $i$-th sample in $\hat{S}$ requires to be calculated, while $\hat{L}_i$ saves how many times the $i$-th sample has passed through the encoder layer.
The main part of this method is described as a while loop in Algorithm 1 from line 6 to line 23, which includes two stages. $\hat{S}$ is filled up in the first stage and used as the input for the second stage. The while loop will transport all the samples from the input dataset to the encoder layer, and after that if the network still hasn't finished all the computation, the last part of the code (starting from line 26) will make this job done.
The latency of the entire model is mainly caused by the encoder, but the stage of data filling also needs to be dealt with carefully. Otherwise some serial operations may be time-consuming when the batch size is increased. For example, the embedding layer (in line 9) should process the input sentence parallelly to speed up the calculation as possible.

Several other facts about this acceleration approach need to be kept in mind. First, this method mainly focuses on high-level optimization, making it have good portability. Because the internal calculation of each encoder is not modified, many other techniques can be compatible with our method for further improvements, such as model distillation, low-bit quantization, or underlying operator optimization. Although this approach can be extended to other BERT-like models, as long as it can apply an early exit strategy and a parameter-sharing strategy, it is highly probable that this approach cannot be applied to a model like CNN. This is because CNNs tend to have different shapes of input tensor and output tensor for each layer, which makes the parameter-sharing strategy hard to be applied. Lastly, how the text data is fed into the model is also critical. This method not only uses parameter compression and early exit on the model, but also has a suitable arrangement of the input data so that ELBERT can keep running until the last input sentence, as shown in case 6 of Fig. 5.

\begin{algorithm}
\caption{ELBERT-Based Acceleration Method}
\label{alg}
\begin{algorithmic}[1]
\STATE \textbf{Input:} The input dataset
\STATE $\hat{S} \gets $ An empty tensor with shape as (N, seq\_len, hidden\_state)
\STATE $\hat{F}_{1,2,...,N} \gets$ [False, False, ... , False]
\STATE $\hat{L}_{1,2,...,N} \gets $ [0, 0, ... , 0]
\STATE Return\_List $\gets$ [0, 1, ... , $N$-1]
\WHILE {Input dataset is not empty}
\STATE $k \gets $ len(Return\_List)
\STATE Input\_Sentence $\gets$ the next $k$ samples from the input dataset
\STATE add Embedding\_Layer(Input\_Sentence) to $\hat{S}$
\STATE $\hat{F}_{1,2,...,N} \gets$ [True, True, ... , True]
\STATE Return\_List $\gets$ empty
\WHILE {$\hat{F}_{1,2,...,N} =$ [True, True, ... , True]}
\STATE $\hat{S} \gets$ Encoder($\hat{S}$)
\STATE $\hat{L}_{1,2,...,N} \gets \hat{L}_{1,2,...,N}+1$
\FOR {$i = 0, 1, ... , N-1$ }
\IF {$\hat{S}_i$ can exit \textbf{OR} $\hat{L}_i = d$}
\STATE $\hat{F}_i \gets$ False
\STATE add $i$ to Return\_List
\STATE $\hat{L}_i \gets 0$
\ENDIF
\ENDFOR
\ENDWHILE
\ENDWHILE
\WHILE {$\hat{F}_{1,2,...,N} \neq$ [False, False, ... , False]}
\FOR {$i = 0, 1, ... , N-1$ }
\IF {$\hat{F}_i =$ False}
\STATE $\hat{S}_i \gets$ An empty tensor with shape as (1, seq\_len, hidden\_state)
\ENDIF
\ENDFOR
\STATE $\hat{S} \gets$ Encoder($\hat{S}$)
\STATE $\hat{L}_{1,2,...,N} \gets \hat{L}_{1,2,...,N}+1$
\FOR {$i = 0, 1, ... , N-1$ }
\IF {$\hat{S}_i$ can exit \textbf{OR} $\hat{L}_i = d$}
\STATE $\hat{F}_i \gets$ False
\STATE $\hat{L}_i \gets 0$
\ENDIF
\ENDFOR
\ENDWHILE
\RETURN The classifier results of all the input samples
\end{algorithmic}
\end{algorithm}

\subsection{The Proposed SK Dataset}
Another problem in FSA is the lack of labeled datasets \cite{mishev2020evaluation}.  To help solve this problem and provide more data for our experiments, we collect data from the SeekingAlpha website \footnote{\footnotesize{https://seekingalpha.com}}, and construct a new FSA dataset labeled by our experts. Seeking Alpha is a very popular investment research website based on freelance contributors' discussions of the financial markets. Founded in 2004, the site is highly regarded, with several million registered users. Thousands of industry analysts serve as contributors to the site, posting commentary and analysis on a variety of companies and stocks over the world.


The SK dataset contains the hotlist financial news on SeekingAlpha from March 2005 to April 2021, covering various aspects such as company performance, international situation, and national policies. We divide all news into three categories: Positive, Negative, and Neutral, based on their likely impact on financial markets. To prevent unbalanced samples during training, we selected 1452 news items. Examples and their distribution can also be seen in Table \ref{RF1} and Table \ref{RF2}.

For more details on our dataset, please refer to our GitHub project.

\begin{table}[htbp]
    \caption{Examples in the SK Dataset}
    \centering
    \begin{threeparttable}
    \begin{tabular}{p{1.1cm}|p{5.3cm}|p{1.1cm}}
    \hline
    \hline
     \textbf{Dataset} &\textbf{News} &\textbf{Sentiment} \\
    \hline
    \hline
    SK&first nine months 2010 company 's net profit rose eur41m eur30m corresponding period 2009&Positive\\
    \hline
    SK&shops located capital region paijat-hame region&Neutral\\
    \hline
    SK&net sales whole fiscal year 2008 lower 2007 operating profit estimated negative&Negative\\
    \hline
    \hline
    \end{tabular}
    \label{RF1}
    \end{threeparttable}
\end{table}

\begin{table}[htbp]
    \caption{Distribution of Sentiment Labels}
    \centering
    \begin{threeparttable}
    \begin{tabular}{p{2cm}|p{1.1cm}|p{1.1cm}|p{1.1cm}|p{1.1cm}}
    \hline
    \hline
    \textbf{SK Dataset}& \textbf{Positive} &\textbf{Negative} &\textbf{Neutral}&\textbf{Summary}\\
    \hline
        \hline
    Training set & 290 & 290 & 290 & 970\\
    \hline
    Validation set & 97 & 97  & 97 & 291\\
    \hline
    Test set & 97 & 97  & 97 & 291\\
    \hline
    Total & 484 & 484  & 484 & 1452\\
    \hline
    \hline
    \end{tabular}
    \label{RF2}
    \end{threeparttable}
\end{table}

\section{Experiments Results of ELBERT}
\label{sec:experiments1}

This section focuses on proving the effectiveness of the CWB early exit mechanism and the ELBERT itself. The experimental results, which are mainly relation curves between the accuracy and the computation cost, show that ELBERT achieves excellent inference acceleration and outperforms other early exit methods used for accelerating BERT. Please note that the computation cost here is a relative value, which should be one if the early exit mechanism is not triggered, so it will not depend on specific computing devices.

Furthermore, to comprehend the CWB early exit mechanism's principles, we also visualize ELBERT's decision-making procedure. As for its ability to boost the text processing speed on GPU and the performance of our entire FSA system, they will be demonstrated in the next section.

\subsection{Baselines}
To demonstrate the advantages of the CWB early exit mechanism, here we choose three baselines:
\begin{itemize}
 \item Original model: ALBERT-large (depth=24) is selected \cite{lan2019albert} in fine-tuning and inference.
 \item Plain compression: We evaluate several smaller-sized models based on ALBERT-large in fine-tuning and inference. Notice that ALBERT has only one encoder, making it possible to set depth manually in inference to achieve different speeds.
  \item Early exit approach: For fair comparison, the early exit methods in DeeBERT \cite{xin2020deebert} and FastBERT \cite{liu2020fastbert} are all applied to ALBERT-large.
\end{itemize}

\subsection{Datasets}
  \label{ssec:task}
In this section, to test the generalization capability of ELBERT, commonly used GLUE benchmark \cite{wang2019glue}, IMDB \cite{imdb}, and AG-news \cite{zhang2015character} are evaluated in the experiments. These three datasets contain diverse NLP tasks such as Natural Language Inference, News Classification, and Sentiment Analysis.

\subsection{Training and Inference Settings}
  \label{ssec:setup}

  \subsubsection{Training} The corresponding hyperparameters are kept the same as \cite{lan2019albert} for GLUE for a fair comparison. For AG-news and IMDB, we employ a learning rate of 3e-5 and a batch size of 32.

  \subsubsection{Inference} 
  To focus on the comparison between different early exit mechanisms, we set the batch size of inference to 1, following prior work\cite{xin2020deebert,teerapittayanon2016branchynet}. The experiments in this section are all done on an NVIDIA 2080Ti GPU.

\subsection{Good Performance of ELBERT}
  \label{ssec:performance}
   \subsubsection{Accelerating inference flexibly}
\begin{figure}[bt]
        \centering
        \includegraphics[width=7cm]{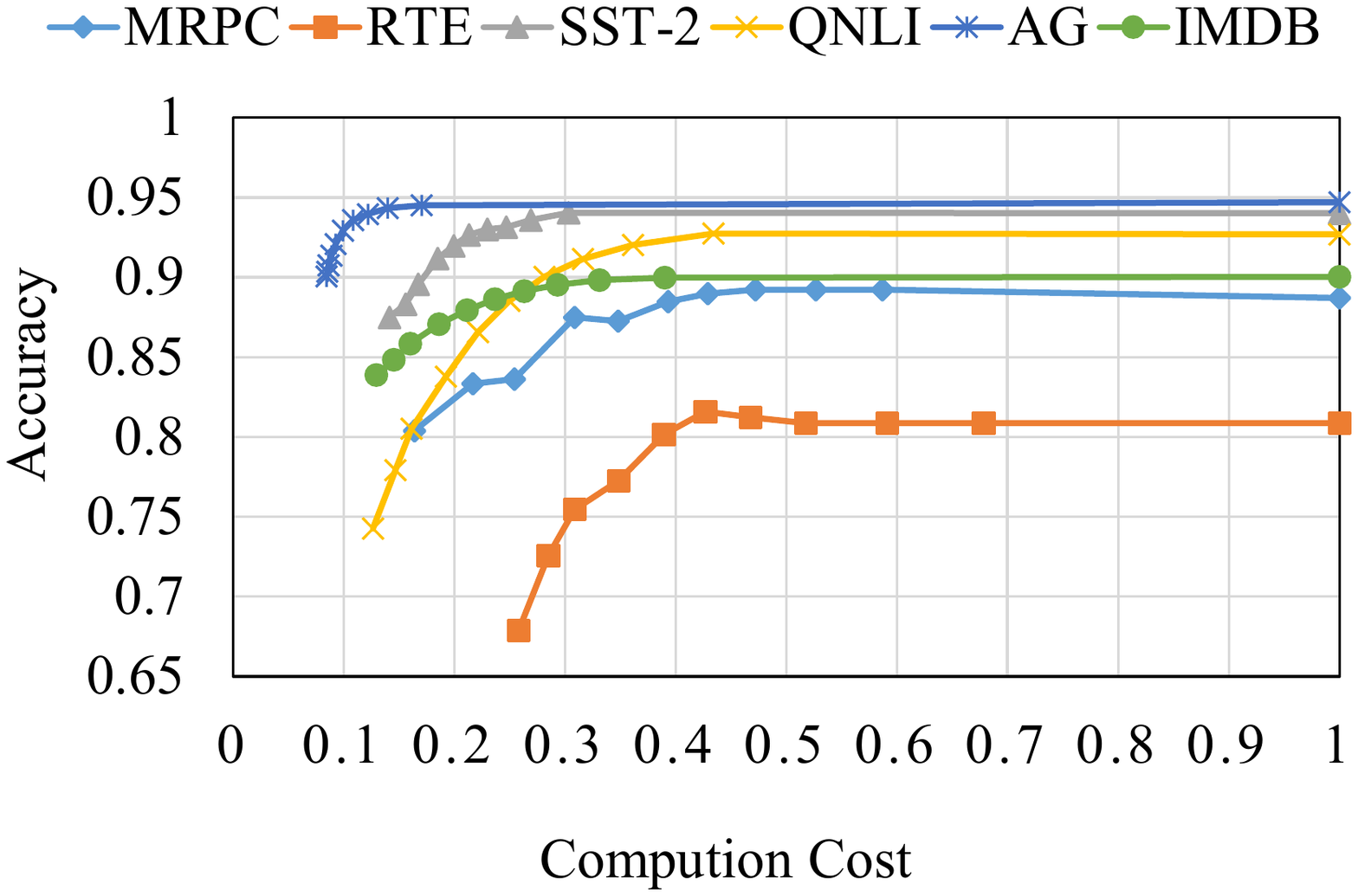}
        \caption{The accuracy-speed tradeoffs of ELBERT on different datasets. The computation cost represents the normalized ratio of the original computation. The rightmost point of each curve denotes the original model without early exit.
        }
        \label{fig:main_compute_ratio_acc}
  \end{figure}

\begin{figure}[!h]
        \centering
        \begin{minipage}[b]{0.25\linewidth}
          \centerline{\includegraphics[width=7cm]{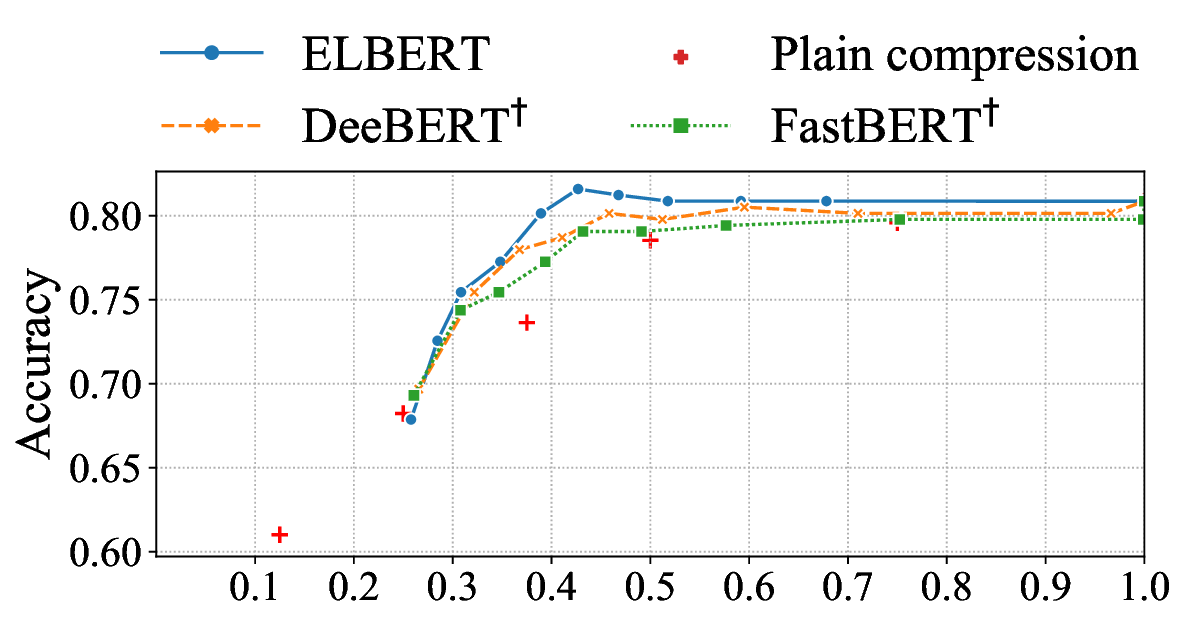}}
        \end{minipage}
        \vfill
        \begin{minipage}[b]{0.25\linewidth}
          \centerline{\includegraphics[width=7cm]{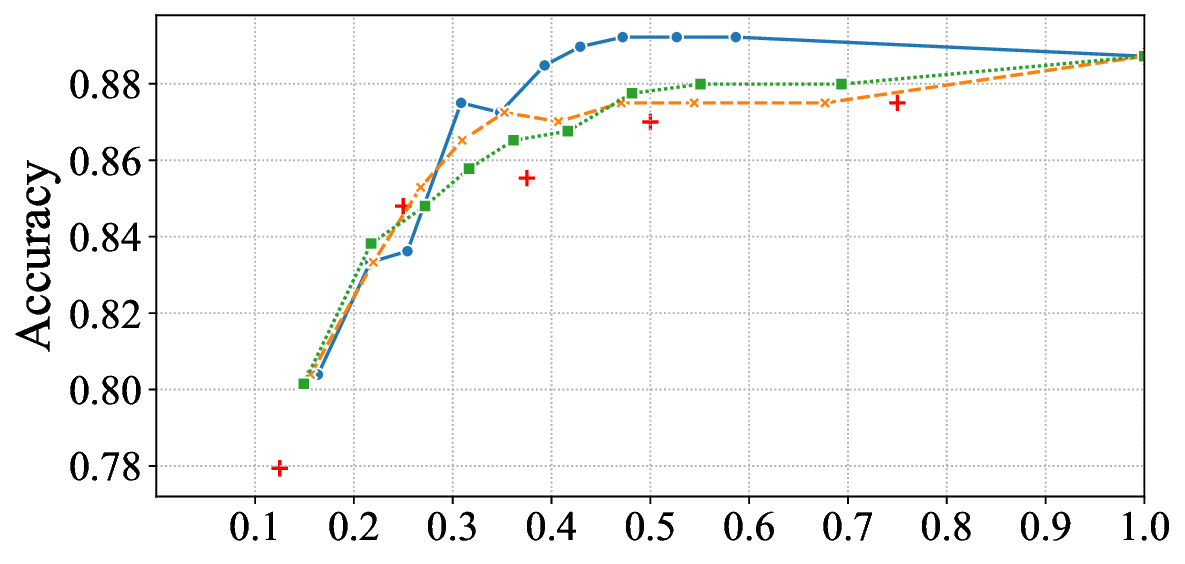}}
        \end{minipage}
        \vfill
        \begin{minipage}[b]{0.25\linewidth}
          \centerline{\includegraphics[width=7cm]{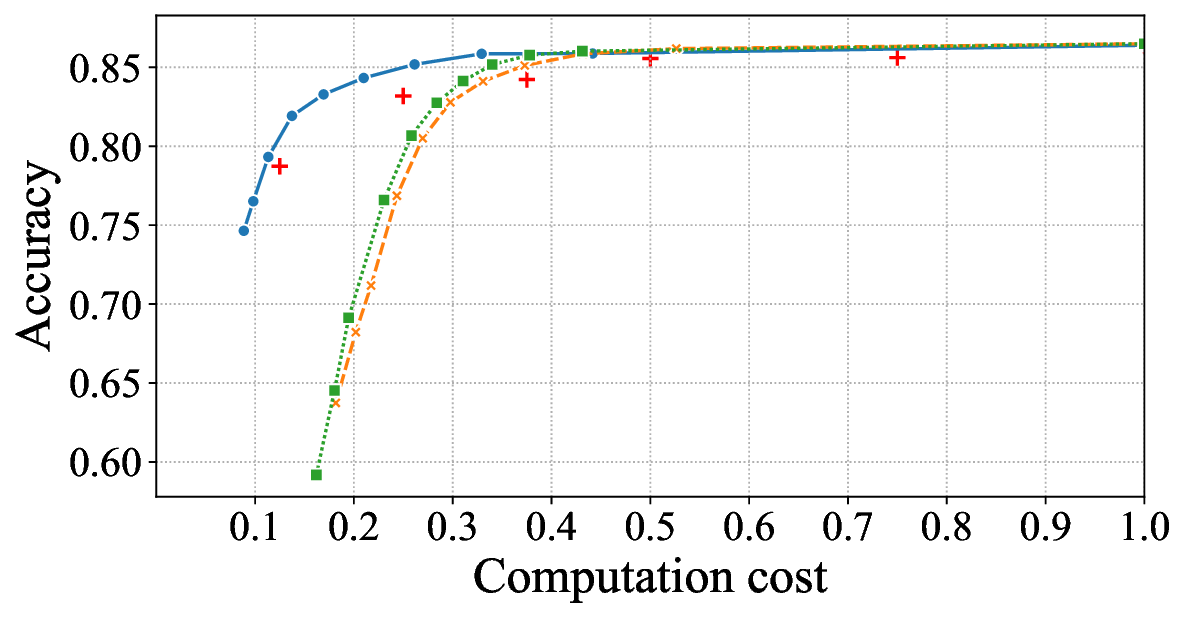}}
        \end{minipage}
        \caption{Comparison on RTE, MRPC and MNLI. Note that the results of DeeBERT and FastBERT are from our implementation on ALBERT-large.
        }
        \label{fig:comparison}
      \end{figure}

   ELBERT is evaluated on the above datasets, and the median of 5 runs are reported in \myfig{fig:main_compute_ratio_acc} and \myfig{fig:comparison}, where the y-axis represents accuracy, and the x-axis represents compute-ratio, the normalized ratio of computation based on FLOPs of samples.
    The curves are drawn by interpolating several points that correspond to different $\delta$, which varies from 0.1 to 1.0 with a step size of 0.1 in the first stage of early exit. ELBERT achieves at least two times inference speedup for all datasets while keeping or even improving the accuracy. The inference acceleration ratio can be up to ten with a tolerable accuracy degradation. These results demonstrate ELBERT's superiority in inference acceleration.

    \subsubsection{Task-related trends} It is worth mentioning that trends of curves in \myfig{fig:main_compute_ratio_acc} are quite different. For example, in the news classification task where ELBERT gets the best performance, the accuracy loss with the smallest computation cost remains very little. In other tasks like the sentiment analysis (SST-2, IMDB) and the natural language inference (QNLI, RTE), when the computation cost is decreased, the curves will drop at a faster speed than the news classification task. This fact indicates that different tasks may have different internal attributes and acceleration difficulties.

\subsubsection{Flexible and better accuracy-speed tradeoffs}
    Comparison results are exhibited in \myfig{fig:comparison}, where the red star-shaped points represent different models obtained by plain compression.
    Our first observation is that ELBERT significantly beats plain compression models. Furthermore, compared to other early-exit-based methods, ELBERT acquires higher accuracy than DeeBERT and FastBERT under the same computation cost, which indicates ELBERT's notable advantages over other approaches.

\subsection{Visualization of CWB}
  \label{ssec:see}
Now we have seen ELBERT's outstanding performance in inference acceleration.
To visualize the decision-making procedure of CWB, we adjust the BertViz\cite{vig2019bertviz}, a tool for visualizing attention in the  Transformer model. The attention scores of each layer are utilized to get the cumulative attention scores, which allows us to see the attention relationships between tokens plainly when the input sample passes through different layers of ELBERT. Since ELBERT only takes the [cls] token as the representation of one sentence to do classification, we only show the cumulative attention scores of [cls] to other tokens in the figures.
Here we take a Sentiment Analysis dataset SST as an example, and two main patterns of early exit are found.

  \subsubsection{Simple input, simple exit} 
\begin{figure}[t]
          \begin{minipage}[b]{0.49\linewidth}
            \centering
            \centerline{\includegraphics[width=4.0cm]{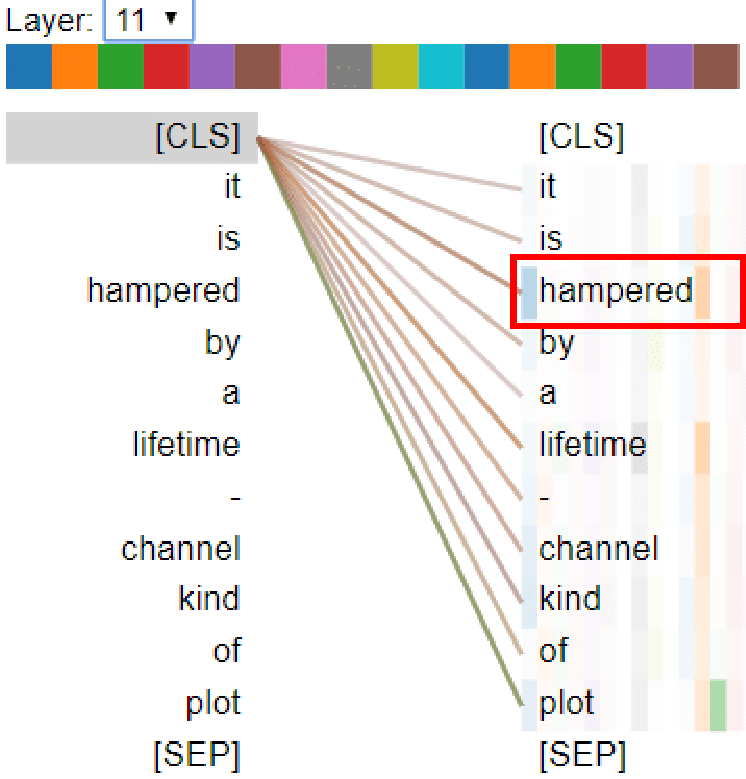}}
            \centerline{(a) Layer 11, $Negative$}\medskip
          \end{minipage}
          \hfill
          \begin{minipage}[b]{0.49\linewidth}
            \centering
            \centerline{\includegraphics[width=4.0cm]{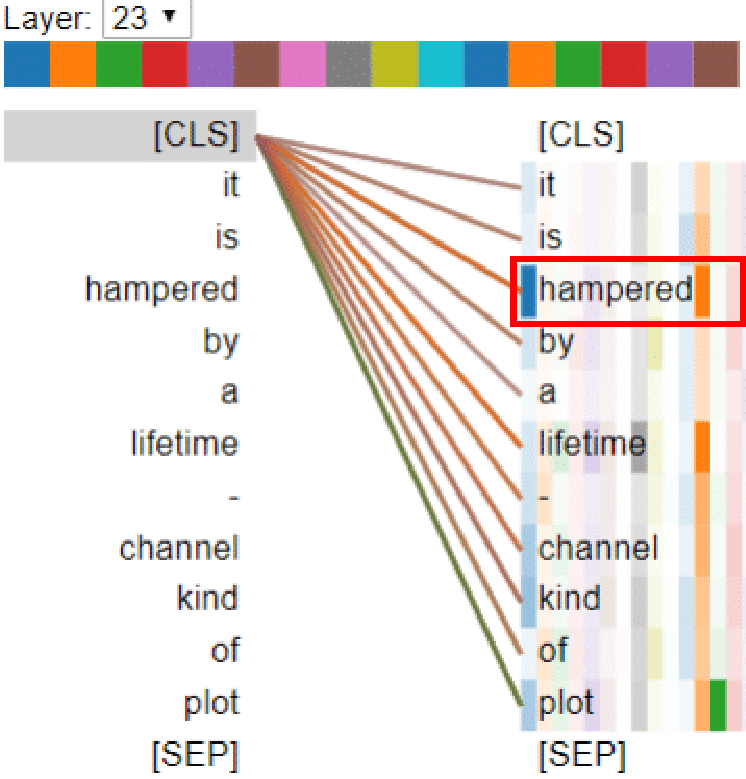}}
            \centerline{(b) Layer 23, $Negative$}\medskip
          \end{minipage}
          \caption{A simple case. The early exit is triggered when the attention to specific word (\textcolor{red}{hampered}) exceeds a certain limit. }
          \label{fig:attn_49}
        \end{figure}

  Generally, if the input sentences have no emotive turns or negatives, the attention of [cls] to the emotional keywords (i.e., hampered) will rise monotonically, which is described in \myfig{fig:attn_49}. Notice that if such attention is larger than a $\delta$-determined limit, the inference of the network will be stopped in advance to reduce redundant computations.

  \subsubsection{Mitigating overthinking}
\begin{figure}[t]
      \begin{minipage}[b]{0.49\linewidth}
        \centering
        \centerline{\includegraphics[width=4.0cm]{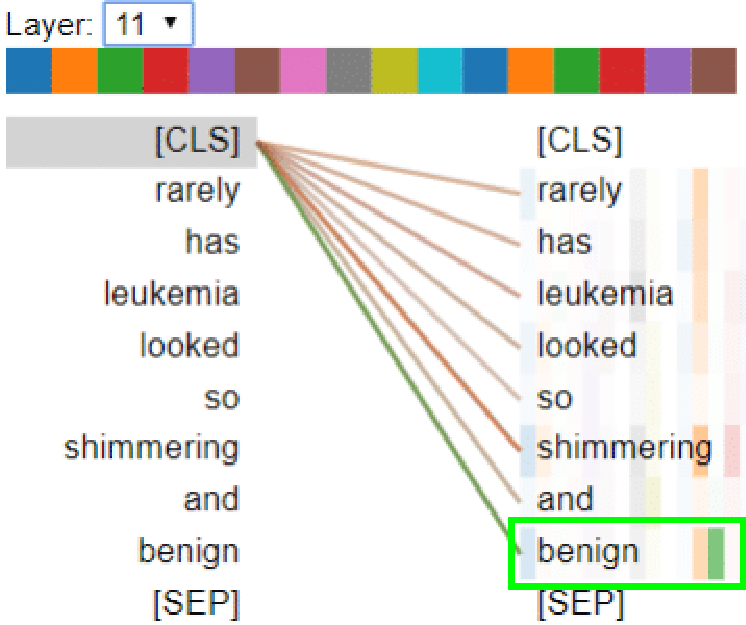}}
        \centerline{(a) Layer 11, $Positive$}\medskip
      \end{minipage}
      \hfill
      \begin{minipage}[b]{0.49\linewidth}
        \centering
        \centerline{\includegraphics[width=4.0cm]{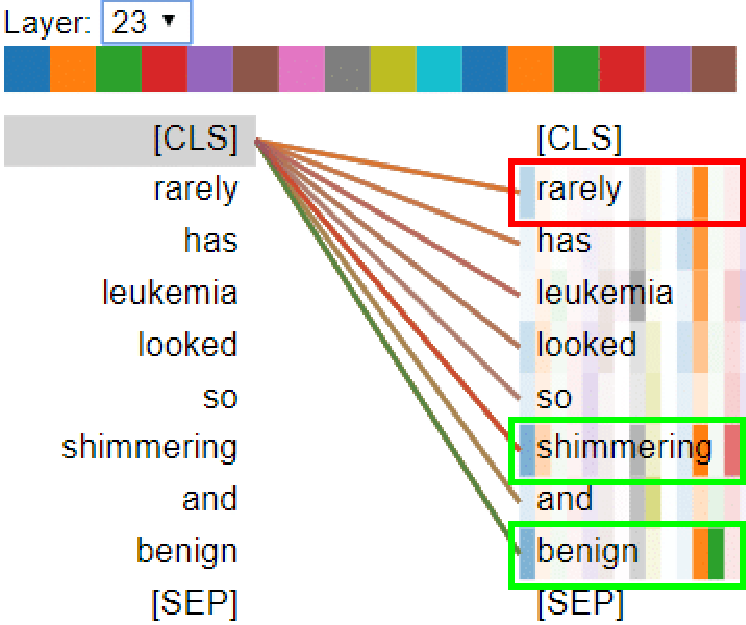}}
        \centerline{(b) Layer 23, $Negative$}\medskip
      \end{minipage}
      \caption{A hard case. The early exit is triggered in time after the commendatory word (\textcolor{green}{benign}) is well noticed, avoiding subsequently overthinking about unrelated negation (\textcolor{red}{rarely}). }
      \label{fig:attn_171}
    \end{figure}

  As is shown in \myfig{fig:comparison}, ELBERT sometimes attains even higher accuracy than that of the original ALBERT-large model, demonstrating that the CWB mechanism fixes some incorrect predictions of the last layer. As shown in \myfig{fig:attn_171}, the model first pays attention to the commendatory word (i.e., benign) and predicts $Positive$. Next, an irrelevant negative word (i.e., rarely) is noticed, seen as the negation of commendatory words. Finally the model forecasts $Negative$, which is a typical example of overthinking.
  In right cases, the negation and the corresponding word are often noticed simultaneously.

  These two patterns illustrate that the CWB mechanism helps establish appropriate attention to some ``keywords'', such as negatives and words with solid emotional orientation. The prediction of ELBERT is strongly affected by these words, and the CWB mechanism enables the network to stop in advance and avoid overthinking.
  


\section{Experiments Results of Our FSA System}
\label{sec:experiments2}

\subsection{The Relationship between Inference Batch Size and Speed on GPU}
\label{gpu batch experiment}
\begin{figure}[bt]
   \centering
        \includegraphics[width=9cm]{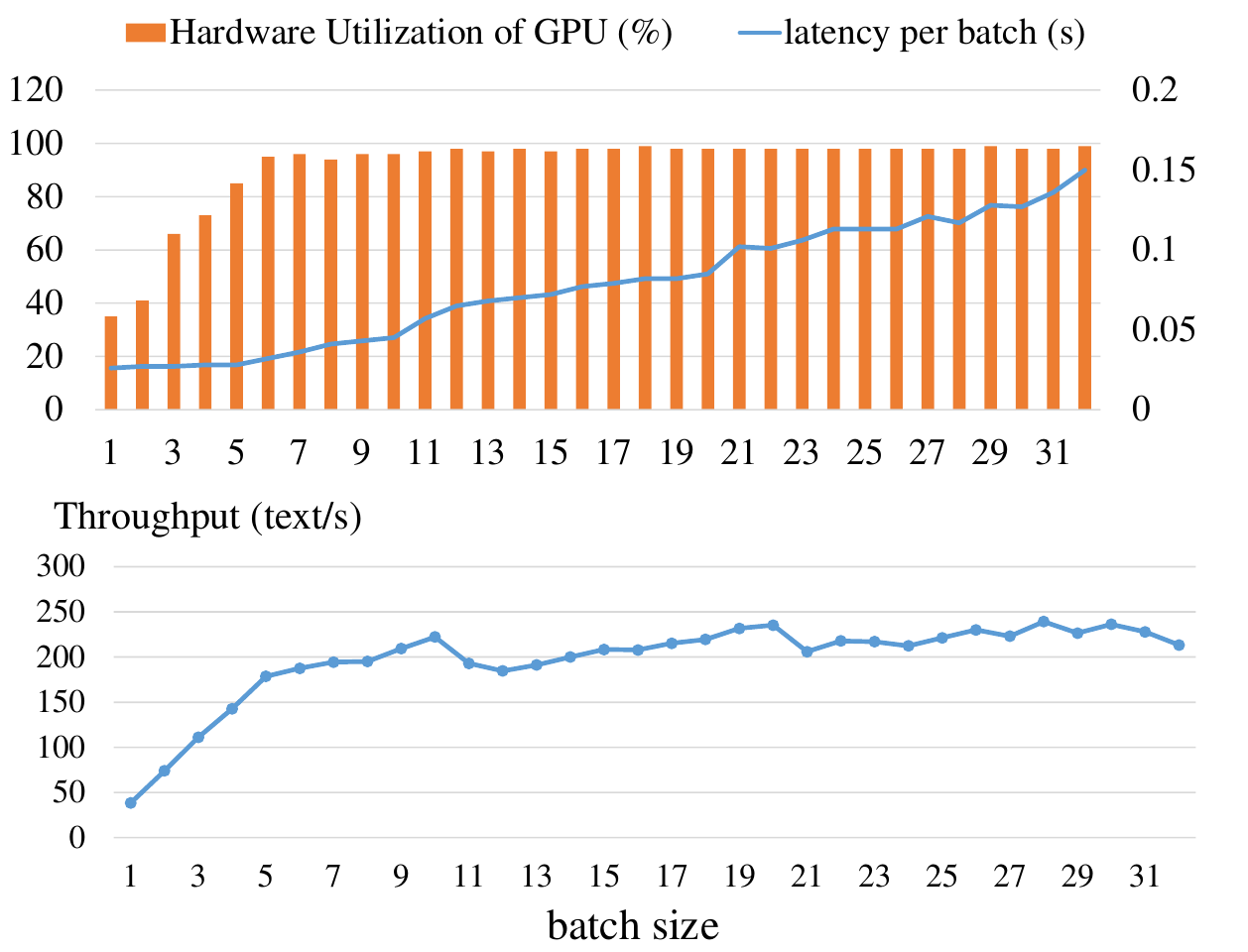}
        \caption{The relationship between the batch size and the processing speed of ALBETR-large when it is deployed on an Nvidia RTX 3090 GPU. The SK dataset is used for test.
        }
        \label{testbs}
  \end{figure}

Before we present the results of the FSA task, we have also carried out some experiments to prove that the inference speed of neural networks can be improved by increasing the batch size of the input, as we mentioned in Section \ref{ssec:accELBERT}.
Some results are shown in Fig. \ref{testbs}, where we use the SK dataset as the input example, and the ALBETR-large is used which is deployed on an Nvidia RTX3090 GPU. In this case, by increasing the batch size, the throughput can be increased from 38 text/s to 240 text/s.

It can be seen that with the expansion of the batch size within certain limits (about 6 or 7), the delay of each batch increases very slowly, which makes the throughput of the entire inference system increase almost linearly with the increase of the batch size.
When the increase of batch size exceeds this limit, the hardware utilization rate will reach a very high value. At this time, increasing the batch size cannot improve the throughput rate as easily. This fact is also consistent with the principle we mentioned in Fig. \ref{fig:ELBERTFAST}: The reason why we can increase the batch size to increase the speed is that the hardware utilization is improved.

In short, the throughput can reach an upper limit by making the batch size larger. Therefore, in the following experiments, we only distinguish between \textbf{batch size being one and batch size being big enough to maximize both the GPU utilization and the processing speed}, and the latter is expressed as ``batch size=N''.

\subsection{Experiment Settings of ELBERT-Based FSA System}
\subsubsection{Baseline Models}
Firstly, FastBERT is used as the main comparison not only because it is one of the latest lightweight BERT models, but also because of its publicly available source code, which supports input batch size greater than one. This feature is very rare among publicly available models, and in general, models that use the early exit method tend to be used only in cases where the batch size is equal to 1. 
Secondly, the DeeBERT model is also chosen, which can only support the input batch size of one. (In the case of ``batch size=N'' in Fig.~\ref{bs1}, the performance of DeeBERT is the same with ``batch size=1''.)
At last, the BERT-large model and the ALBERT-large model are also fine-tuned on our FSA datasets for comparison.
Since here we need to compare the network performance in the overall system, we use the entire FastBERT and DeeBERT rather than only their early exit approaches compared with experiments in Section \ref{sec:experiments1}.
These two light BERTs along with ELBERT are all compressed from BERT-large in the following experiments.

\subsubsection{Our Model}
We use the ELBERT-based acceleration method which is described in Algorithm 1 for the FSA task. During the testing (inference) phase, batch sizes from 1 to 32 are tested, and \textbf{the highest throughput is chosen as the case of ``batch size=N''} like we mentioned before. All the inference experiments are done on a server with an Nvidia RTX 3090 GPU.


\subsubsection{Datasets}
The Financial Phrase-Bank (FPB) dataset \cite{malo2014good} and the proposed SK dataset are both used to fine-tune and test our FSA system. The FPB dataset consists of 4845 financial texts which have already been annotated by financial experts. 80\% of all sentences are set as training examples, and the remaining are set as validation examples.

\subsubsection{Fine-Tune}
All of these training hyperparameters are adjusted to the appropriate value. The learning rate used by FastBERT is 2e-5, and other models use 3e-5.
The training batch sizes are all set to 32.
All these models take ten epochs to finish the fine-tuning, and the FastBERT requires another ten epochs for self-distillation.

\subsection{Results of ELBERT for FSA Task}

We use ELBERT and the proposed ELBERT-based acceleration mechanism to speed up the FSA task. Here we give and analyze the results through horizontal and longitudinal contrast. 
Note that we set a ``Accuracy with tolerable loss'' line for each dataset in Fig. 11. It can be seen that a smaller accuracy loss above the line will give a larger speed gain, while the accuracy loss is almost out of control below the line. We also used this as a yardstick in the comparisons in TABLE III and TABLE IV.

\begin{figure}[bt]
   \centering
        \includegraphics[width=8.8cm]{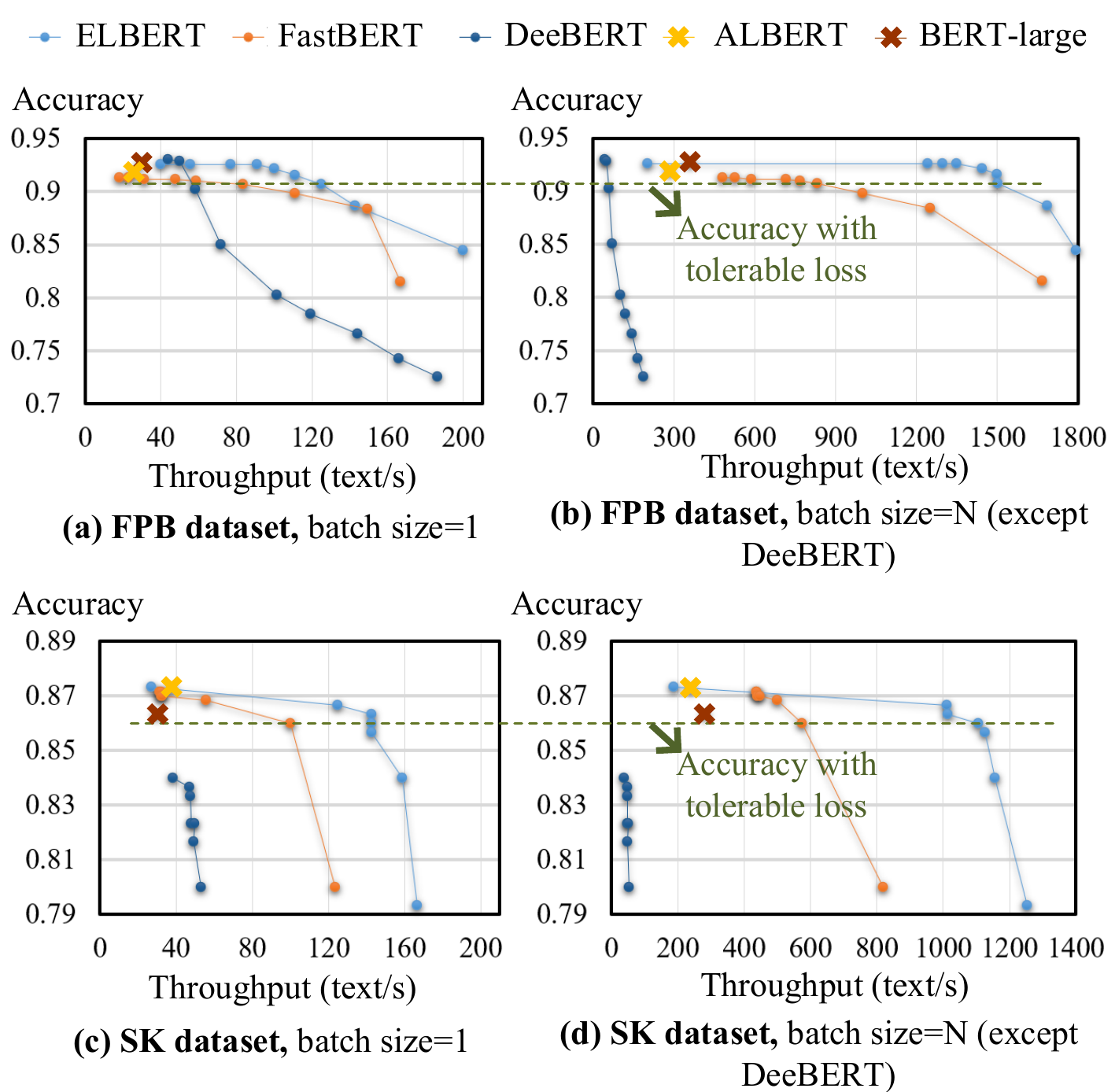}
        \caption{Comparison between the ELBERT and baseline models applied in FSA tasks. Notice that the accuracy will not change when the batch size is increased. Comapred with FastBERT, the best baseline model, in ``batch size=N'' situations ELBERT's lead in speed significantly expands, which also shows the improvement brought by our acceleration method. The points above the dotted lines have acceptable accuracy loss, and ``batch size=N'' (1$<$N$\leq$32) means the batch size is large enough to enable the processing speed to reach an upper limit. These results demonstrate that our method beats all the baseline models, including the original BERT and other light BERTs.
        }
        \label{bs1}
  \end{figure}

On the one hand, the accuracy and throughput results given in Fig.~\ref{bs1} demonstrate that ELBERT significantly outperforms all the baselines in this FSA task. Notice that the original BERT and ALBERT models have no early exit mechanism, so unlike other models, the performance of these two models is represented in the graph as individual points rather than as curves connected by multiple points. 
By decreasing the exit threshold, ELBERT can easily achieve a much higher speed than these two baseline models with very close accuracy. 
Besides, although the DeeBERT model can be accelerated by an early-exit mechanism, it cannot support a batch size larger than one (in Fig. 11 all the results of DeeBERT are measured with batch size one).
Even when the batch size is one, DeeBERT is clearly inferior to ELBERT and FastBERT.

The best baseline model is the FastBERT, which can also support batch sizes larger than 1 while using an early exit mechanism.
However, the ELBERT still has a much better performance compared with FastBERT. 
In the case of bath size one, the curves of ELBERT are closer to the top right corner, which indicates that ELBERT can calculate faster with the same accuracy, as well as ELBERT's higher accuracy with the same speed. In the case of batch size N, this gap is further widened, and with sufficient accuracy, our method is nearly twice as fast as FastBERT. 
Pay attention to the intersections of the model accuracy-throughput curves and the two``Accuracy with tolerable loss'' lines, where the best trade-off is achieved: The inference speed is improved as much as possible while the accuracy loss is acceptable. The horizontal coordinates (throughput rates) of these intersections of the curves (ELBERT and FastBERT) are organized in TABLE \ref{speedcomp}. It can be seen that ELBERT is only 1.4$\sim$1.5 times faster than FastBERT when the batch size is one. Furthermore, when the batch size is N, ELBERT is 1.8$\sim$1.9 times faster than FastBERT, illustrating that our method can improve the efficiency of the early exit method. In contrast, encoder layers of FastBERT have different parameters, which makes it unable to utilize the acceleration method we proposed in Section \ref{section4}.



On the other hand, the results of ELBERT itself are also enough to prove that the ELBERT-based acceleration method can make good use of both the early exit mechanism and higher parallelism. The speedup achieved by our method is calculated in TABLE \ref{speedup}. In experiments using the FPB dataset, the early exit first achieves 3.13$\times$ speed up with acceptable accuracy loss, and by increasing the batch size, another 12$\times$ is achieved. On the SK dataset, these two numbers are 5.26 and 7.94, respectively. These results prove that our method can make the early exit mechanism more effective when the input batch size is set to a big value.

\begin{table}[!ht]
    \caption{Speed Comparison between the ELBERT and FastBERT with Tolerable Accuracy Loss}
    \centering
    \begin{tabular}{c|c|c|c|c|c}
    \hline
    \hline
        \multirow{2}*{Datasets} & \multirow{2}*{Accuracy} &  \multirow{2}*{Batch size} & \multicolumn{3}{c}{Speed (text/s)} \\ \cline{4-6}
        ~ & ~ & ~ & FastBERT & ELBERT & Ratio \\ 
        \hline
        \hline
        \multirow{2}*{FPB} & \multirow{2}*{0.907} & 1 & 83 & 125 & 1.50x \\ \cline{3-6}
        ~ & ~ & N & 833 & 1503 & \textbf{1.80x} \\ \hline
        \multirow{2}*{SK} & \multirow{2}*{0.860} & 1 & 100 & 143 & 1.43x \\ \cline{3-6}
        ~ & ~ & N & 574 & 1107 & \textbf{1.93x} \\ 
    \hline
    \hline
    \end{tabular}
    \label{speedcomp}
\end{table}

\begin{table}[htbp]
    \caption{The Speedup Achieved by the ELBERT-Based Acceleration Method in the FSA Task}
    \centering
    \begin{threeparttable}
    \begin{tabular}{p{1.9cm}|p{1.2cm}|p{1.2cm}|p{2.4cm}}
    \hline
    \hline
    \textbf{FPB Dataset}\\
    \hline
    Batch size &1 &1 &N \\
    \hline
    Early eixt &no &yes &yes\\
    \hline
    Accuracy & 0.926 & 0.907 & 0.907 \\
    \hline
    Speed (text/s) & 40 & 125  & 1503 \\
    \hline
    Speed-up & 1$\times$ & 3.13$\times$  & 37.6$\times$ (3.13$\times$12.0)\\
    \hline
    \hline
    \textbf{SK Dataset}\\
    \hline
    Batch size &1 &1 &N \\
    \hline
    Early eixt &no &yes &yes\\
    \hline
    Accuracy & 0.873 & 0.860 & 0.860 \\
    \hline
    Speed (text/s) & 27	 & 143	  & 1107 \\
    \hline
    Speed-up & 1$\times$ & 5.29$\times$  & 41.0$\times$ (5.29$\times$7.74)\\
    \hline
    \hline
    \end{tabular}
    \label{speedup}
    \end{threeparttable}
\end{table}

\section{Conclusion}
\label{conclusion}
This paper introduces a compressed BERT model and demonstrates its great potential in NLP tasks desiring both high speed and high accuracy such as FSA.

Firstly, the ELBERT and a CWB early exit mechanism are proposed.
The ELBERT has fewer parameters (18M) and faster inference speed (2$\sim$10$\times$ speedup on various datasets), and on many NLP tasks, the CWB mechanism also outperforms existing early exit methods used for accelerating BERT. Based on the ELBERT, a novel acceleration method is developed to creatively solve the incompatibility between early exit and a high parallelism on GPU. This method also helps us build a fast and high-accuracy FSA system.
With sufficient accuracy and higher throughput, the FSA system can label more text data in the same time or finish processing the same input samples with lower latency. ELBERT has achieved an accuracy of 0.907 with a throughput of 1503 text/s on the FPB dataset. The accuracy and throughput of the SK dataset are 0.860 and 1107 text/s, respectively. Compared with other compressed BERTs, the ELBERT also has much better accuracy and speed in the FSA task.

Since the ELBERT model and the acceleration method are inspiring and have good portability, we will try to bring them into more NLP applications in future work. Besides, we will also attempt to make our FSA system function better for quantitative or event-driven investment strategies.


\bibliographystyle{IEEEtran}
\bibliography{bare_jrnl_winedt2}

\begin{IEEEbiography}[{\includegraphics[width=1.1in,height=1.1in,clip,keepaspectratio]{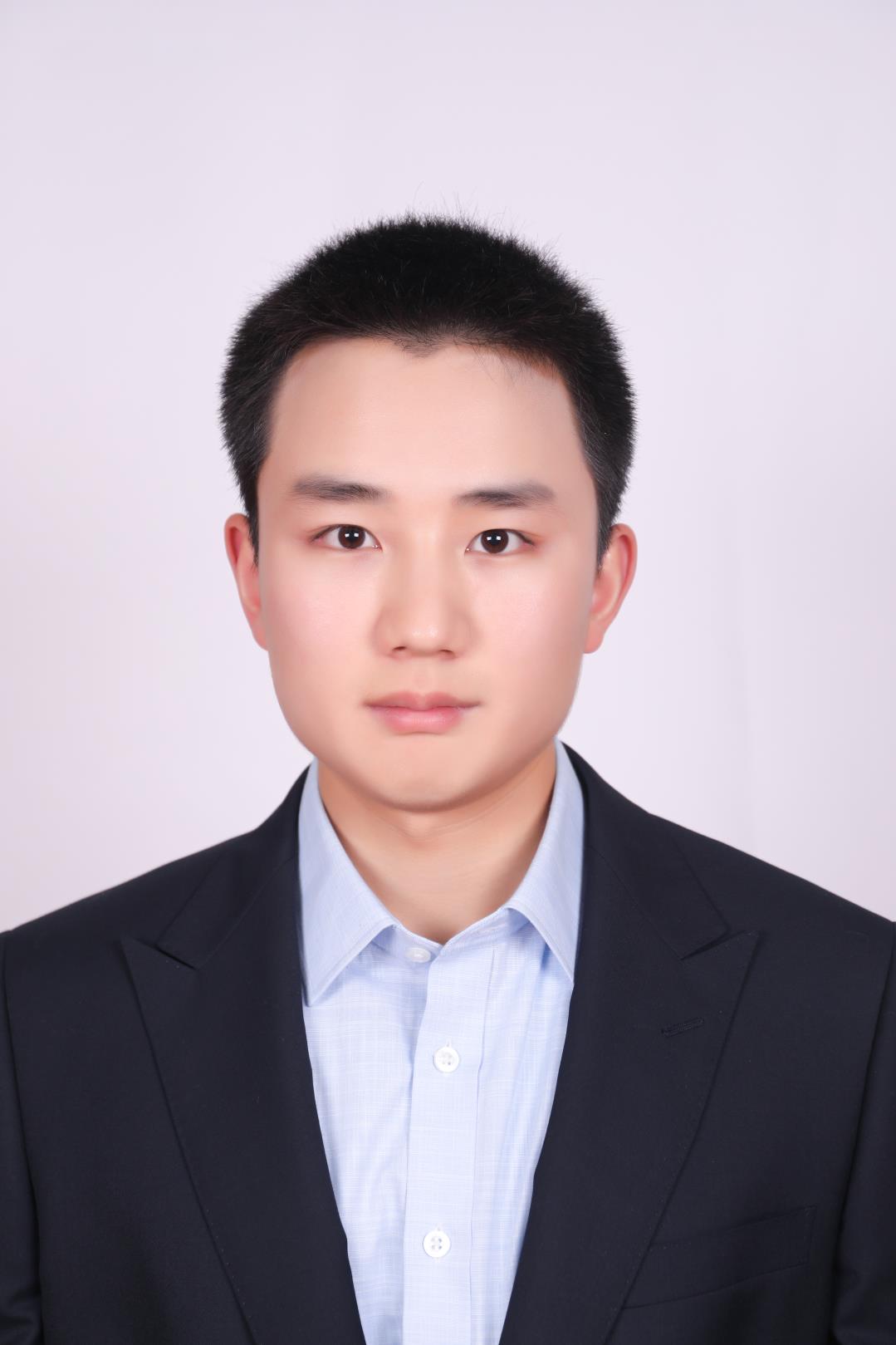}}]{Siyuan Lu}
received the B.S. degree in communication engineering from Nanjing University, Nanjing, China.
Now he is currently pursuing the Ph.D. degree
in information and communication engineering with Nanjing University, Nanjing, China.

His research interests include algorithm optimization and VLSI design for deep learning related applications, especially natural language processing tasks. As the first author of "Hardware Accelerator for Multi-Head Attention and Position-Wise Feed-Forward in the Transformer", he received the Best Paper Award at the IEEE International Systems-on-Chip Conference (SOCC) in 2020.
\vspace{-0.4in}
\end{IEEEbiography}

\begin{IEEEbiography}[{\includegraphics[width=1in,height=1in,clip,keepaspectratio]{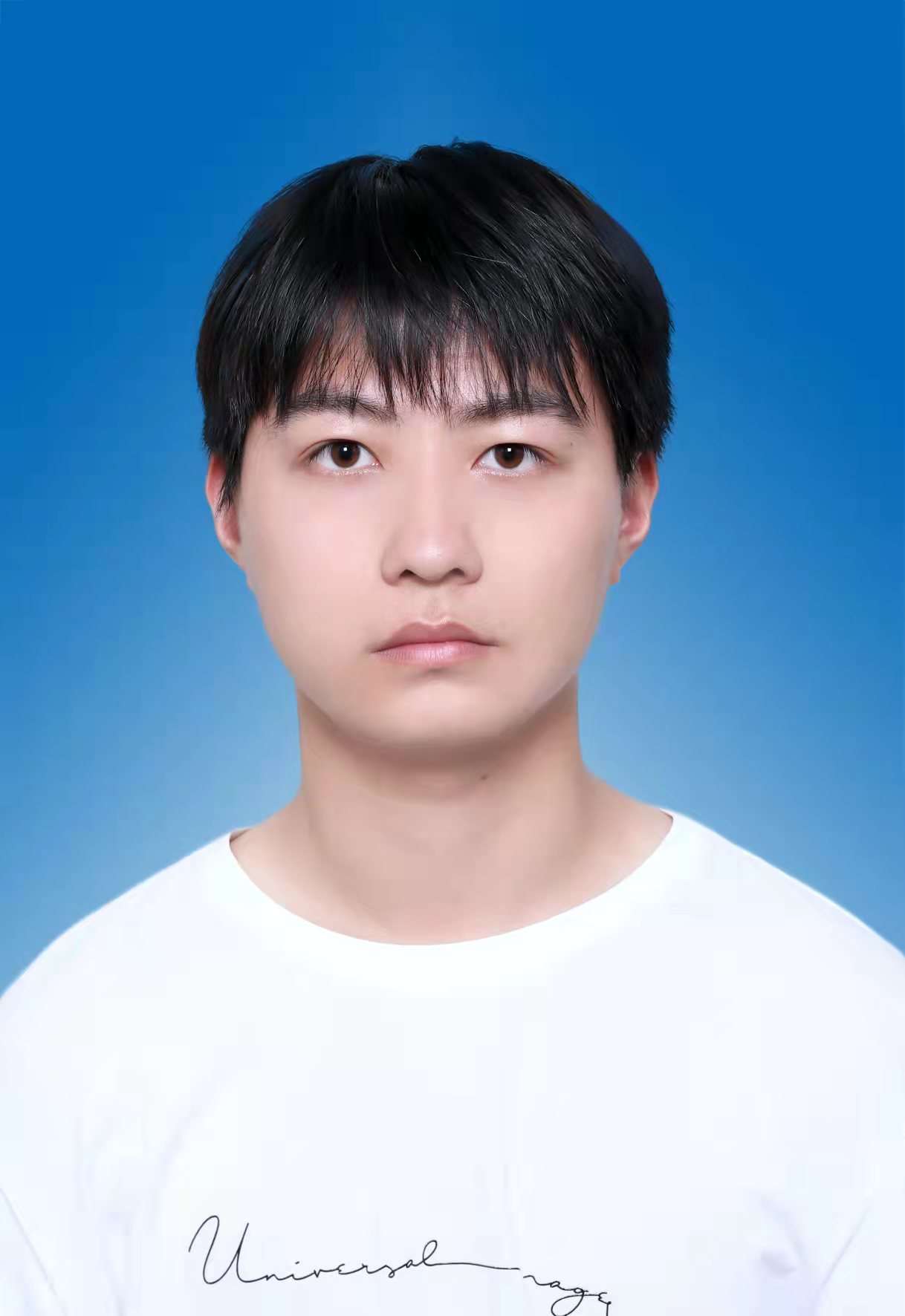}}]{Chenchen Zhou}
received the B.S. degree in electronic information science and technology from Nanjing University, Nanjing, China, where he is currently pursuing the master 's degree in integrated circuit engineering.
His research interests include natural language processing and quantitative trading, especially applying pre-trained language models to trading strategies.
\vspace{-0.4in}
\end{IEEEbiography}

\begin{IEEEbiography}[{\includegraphics[width=1in,height=1in,clip,keepaspectratio]{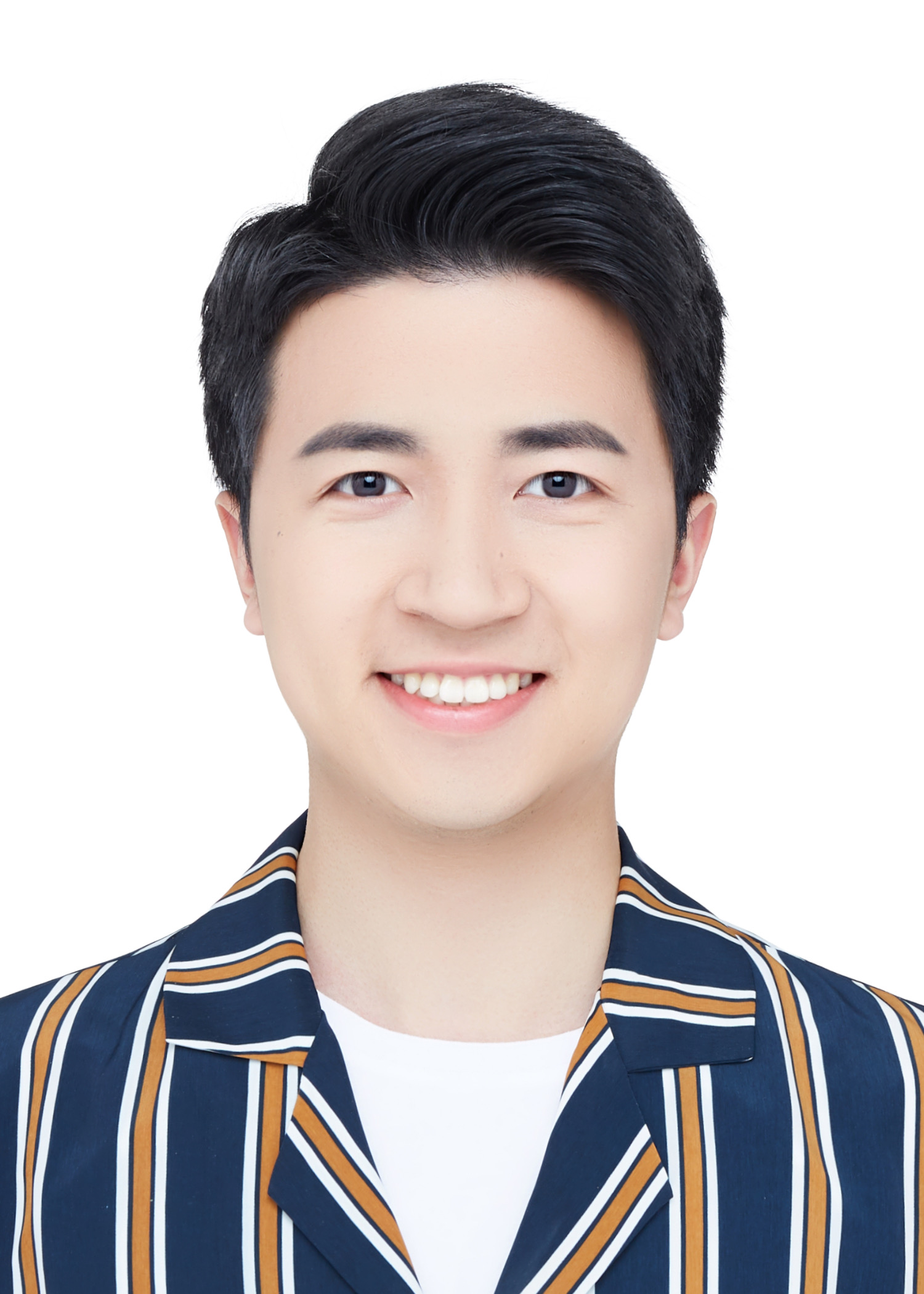}}]{Keli Xie}
received the B.S.degree in integrated circuit design and integrated system from Nanjing University, Nanjing, China, where he is currently pursuing the master's degree in integrated circuit design. His current research interests include efficient hardware for AI, model compression and text summarization, especially dialogue summarization.
\vspace{-0.4in}
\end{IEEEbiography}

\begin{IEEEbiography}[{\includegraphics[width=0.9in,height=1in,clip,keepaspectratio]{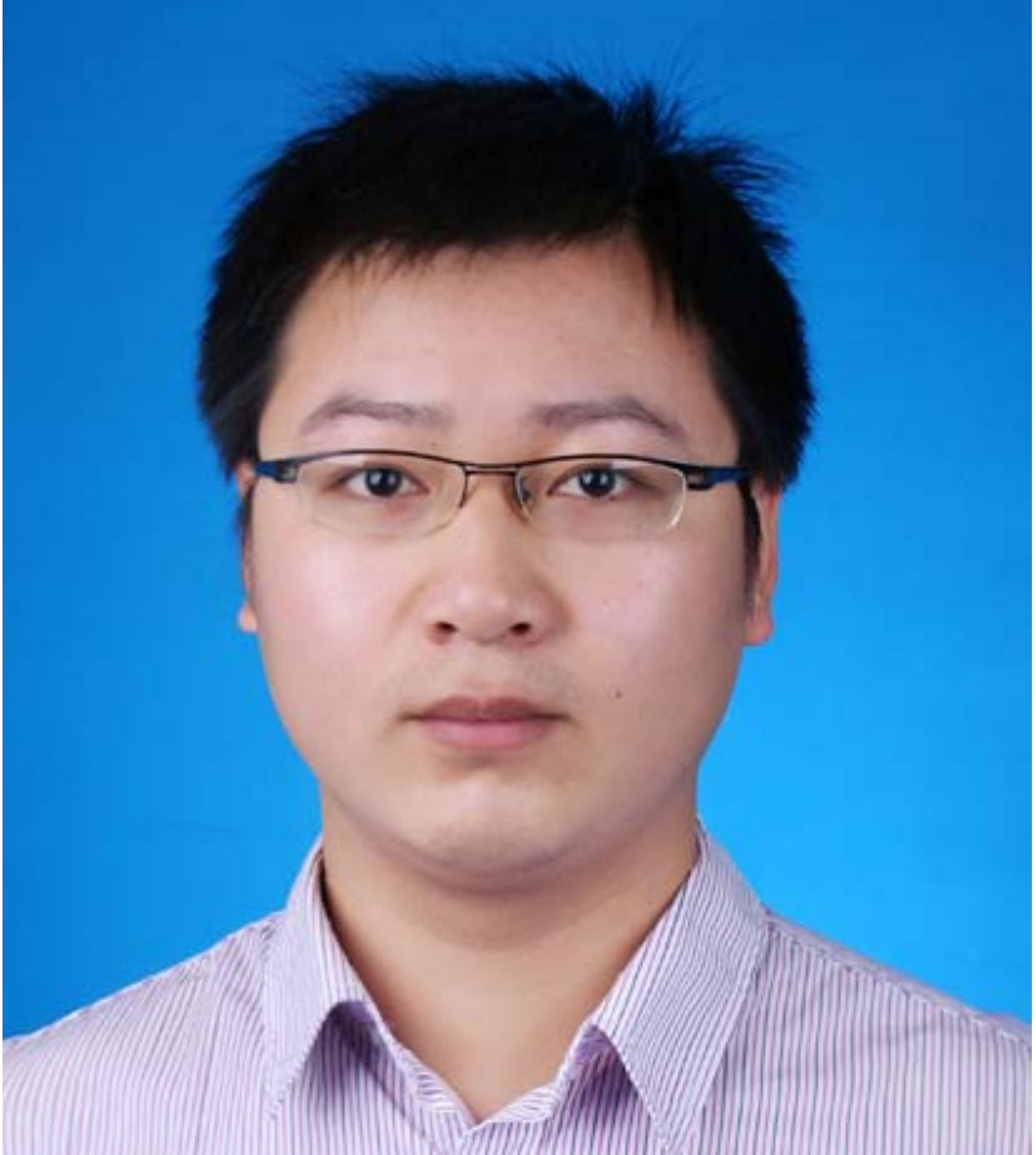}}]{Jun Lin}
	
	(S¡¯14-SM¡¯18) received the B.S. degree in physics and the M.S. degree in microelectronics  from Nanjing University, Nanjing, China, in 2007 and 2010, respectively, and the Ph.D. degree in electrical engineering from the Lehigh University,  Bethlehem, in 2015. From 2010 to 2011, he was an ASIC design engineer with AMD. During summer 2013, he was an intern with Qualcomm Research, Bridgewater, NJ. In June 2015, he joined the school of electronic science and engineering of Nanjing University, where he is an associate professor. He was a member of the Design and Implementation of Signal Processing Systems (DISPS) Technical Committee of the IEEE Signal Processing Society. His current research interests include low-power high-speed VLSI design for digital signal processing and deep learning, hardware acceleration for big data processing and emerging computer architectures. He was a co-recipient of the Best Paper Award at the IEEE Computer Society Annual Symposium on VLSI (ISVLSI) in 2019, the Best Paper Award (The First Place) at the  IEEE International Signal Processing Systems (SiPS) in 2019, the Merit Student Paper Award at the IEEE Asia Pacific Conference on Circuits and Systems in 2008. He was a recipient of the 2014 IEEE Circuits \& Systems Society (CAS) student travel award.
	\vspace{-0.4in}
\end{IEEEbiography}

\begin{IEEEbiography}[{\includegraphics[width=1in,height=1.1in,clip,keepaspectratio]{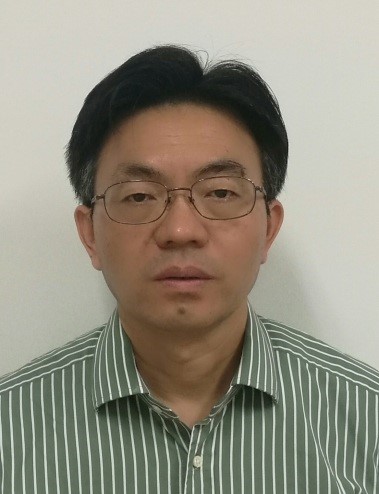}}]{Zhongfeng Wang}
	received both the B.E. and M.S. degrees in the Dept. of Automation at Tsinghua University, Beijing, China, in 1988 and 1990, respectively. He obtained the Ph.D. degree from the University of Minnesota, Minneapolis, in 2000. He has been working for Nanjing University, China, as a Distinguished Professor since 2016. Previously he worked for Broadcom Corporation, California, from 2007 to 2016 as a leading VLSI architect. Before that, he worked for Oregon State University and National Semiconductor Corporation.

Dr. Wang is a world-recognized expert on Low-Power High-Speed VLSI Design for Signal Processing Systems. He has published over 200 technical papers with multiple best paper awards received from the IEEE technical societies, among which is the VLSI Transactions Best Paper Award of 2007. He has edited one book VLSI and held more than 20 U.S. and China patents. In the current record, he has had many papers ranking among top 25 most (annually) downloaded manuscripts in IEEE Trans. on VLSI Systems. In the past, he has served as Associate Editor for IEEE Trans. on TCAS-I, T-CAS-II, and T-VLSI for many terms. He has also served as TPC member and various chairs for tens of international conferences. Moreover, he has contributed significantly to the industrial standards. So far, his technical proposals have been adopted by more than fifteen international networking standards. In 2015, he was elevated to the Fellow of IEEE for contributions to VLSI design and implementation of FEC coding. His current research interests are in the area of Optimized VLSI Design for Digital Communications and Deep Learning.

\end{IEEEbiography}

\end{document}